\documentclass{article}
\usepackage{enumitem}

\usepackage[nonatbib, preprint]{neurips_2026}

\usepackage[numbers,sort&compress]{natbib}
\usepackage[utf8]{inputenc} % allow utf-8 input
\usepackage[T1]{fontenc}    % use 8-bit T1 fonts
\usepackage{hyperref}       % hyperlinks
\usepackage{url}            % simple URL typesetting
\usepackage{booktabs}       % professional-quality tables
\usepackage{amsfonts}       % blackboard math symbols
\usepackage{nicefrac}       % compact symbols for 1/2, etc.
\usepackage{microtype}      % microtypography
\usepackage{xcolor}         % colors
\usepackage{graphicx}
\usepackage{amsmath}
\usepackage{subcaption}
\usepackage{array}
\usepackage{placeins}

\title{Capability $\neq$ Interpretability: Human Interpretability of Vision Foundation Models}

\author{%
  Julien Colin \\
  ELLIS Alicante, Spain\\
  Brown University, USA\\
  \texttt{julien$\_$colin@brown.edu} \\
  \And
  Lore Goetschalckx \\
  imec, Leuven, Belgium \\
  \AND
  Thomas Serre\thanks{co-principal investigators} \\
  Brown University, USA \\
  \And
  Nuria Oliver$^{*}$ \\
  ELLIS Alicante, Spain \\
}

\begin{document}

\maketitle

\begin{abstract}
How interpretable are the features of leading vision models? The question is increasingly pressing as these models move from research benchmarks into high-stakes deployments, yet existing methods cannot answer it reliably. 
We address this gap with a framework for measuring and comparing the human interpretability of vision models, built around two complementary psychophysics protocols: (1) \emph{localizability}---can an observer predict \emph{where} a feature fires on a novel image?---and (2) \emph{semanticity}---can an observer accurately describe \emph{what} the feature represents? Features are recovered via sparse autoencoders, and a chance-anchored scoring function places every model on a common scale. Applying the framework to six vision transformers---two supervised ViTs and four foundation models (DINOv2, DINOv3, CLIP, SigLIP)---we collected more than $15{,}000$ behavioral responses, of which we analyzed $13{,}400$ responses from the $377$ participants who passed our pre-specified quality checks. We find that foundation models are consistently \emph{less} interpretable than their size-matched supervised counterparts, and the gap is not a capability tradeoff: interpretability does not correlate with downstream task performance on any benchmark we examine. What does correlate is the \emph{locality} of a feature's activations and \emph{coarse-grained} semantic alignment with humans---models with focal activations and representations that reflect the world's broad categorical structure produce more interpretable features, whereas fine-grained perceptual alignment does not. The two protocols yield strongly correlated rankings and share the same predictors, establishing interpretability as an independent, measurable dimension of representation quality---and, surprisingly, one on which the foundation models we tested generally fall below their size-matched supervised baseline. Capability alone cannot close that gap, but locality and coarse-grained alignment might.
\end{abstract}

\section{Introduction}

The dominant paradigm in modern computer vision is to pretrain large foundation models and fine-tune them for specific tasks. Built on the vision transformer~\citep{dosovitskiy2021vit}, models like CLIP~\citep{radford2021learning}, SigLIP~\citep{zhai2023sigmoid}, DINOv2~\citep{oquab2023dinov2}, and DINOv3~\citep{simeoni2025dinov3} already serve as general-purpose visual backbones across an ever-expanding range of downstream tasks, including classification~\cite{chiu2024fine}, segmentation~\cite{liu2024groundingdinomarryingdino}, object tracking~\cite{faber2024leveraging, tumanyan2024dino}, and robotic perception~\cite{kim2025openvla}. The same models are increasingly deployed in high-stakes settings---clinical decision support~\cite{baharoon2023dinoradio, moutakanni2024dinoX, foundationmedical, d2025foundationradio}, autonomous driving~\cite{gao2025fsurvey, jiang2025fsurvey}---where a user often needs to understand what a model is actually responding to. How interpretable, then, are the features that today's leading vision models have learned?

Surprisingly little work compares vision models on this dimension, because existing evaluation protocols cannot place different models on a common, reliable scale of feature interpretability (Section~\ref{sec:related}).
\begin{figure}[t!]
  \centering
\includegraphics[width=\textwidth]{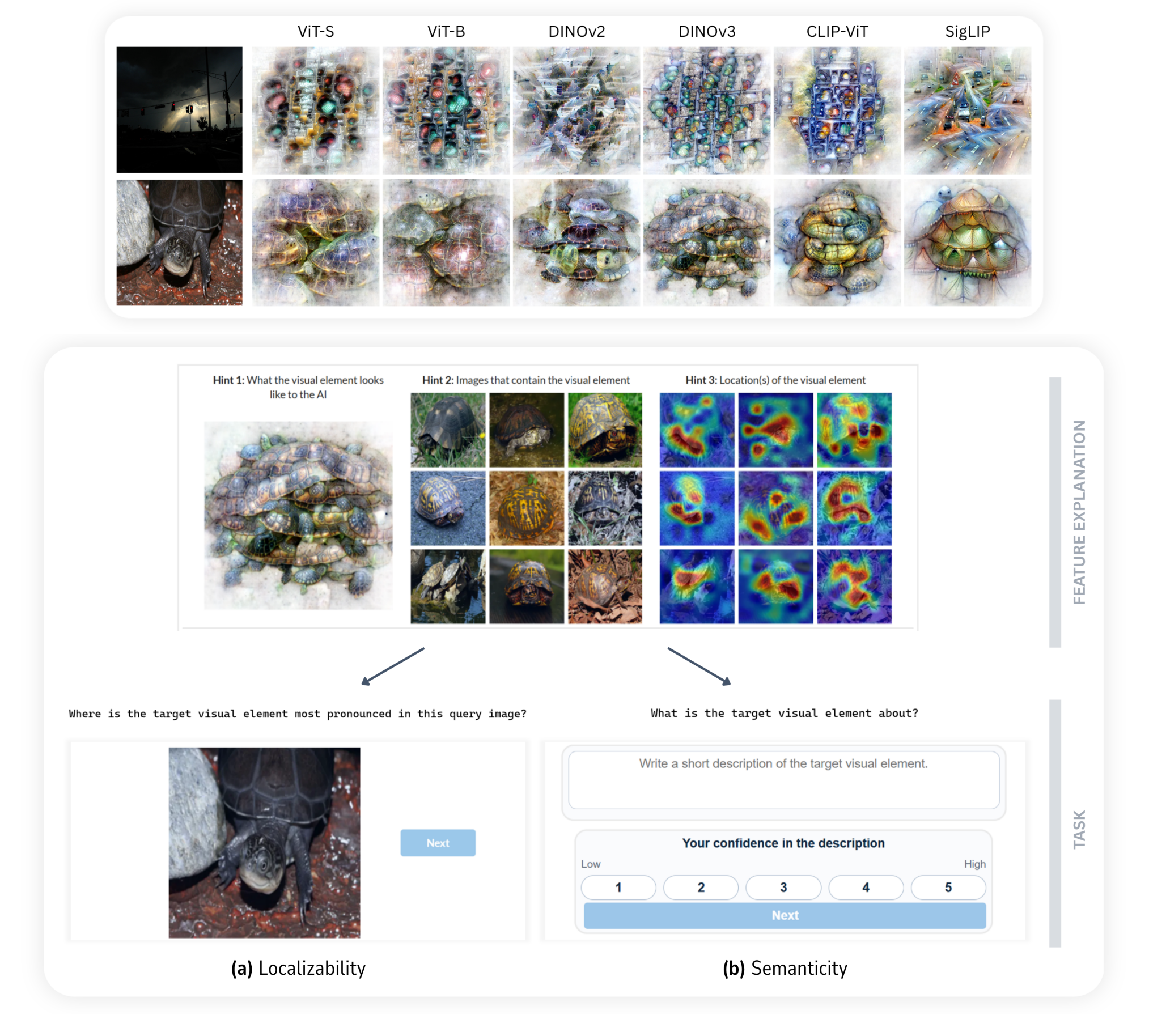}
  \caption{Models trained under different objectives learn different internal representations and, consequently, different features. Which of these features are more interpretable to humans remains an open question. \textbf{Top:} the most important feature recovered by each of the six models we study, for the prediction of \textit{traffic light} (first row) and \textit{mud turtle} (second row). \textbf{Bottom:} Our framework, illustrated for DINOv3. Given a feature and its visual explanation---a synthetic visualization, highly activating images, and corresponding heatmaps showing where the feature fires (top panels)---participants either click where they think the feature fires on a novel image (left, \emph{localizability}) or describe what the feature represents in free text (right, \emph{semanticity}).}
\label{fig:interface}\label{fig:features}
\end{figure}

To fill this gap, we introduce a model evaluation framework comprising two psychophysics protocols that quantify two complementary dimensions of feature understanding: feature \emph{localizability} — can an observer predict \emph{where} a feature fires on a novel image? — and feature \emph{semanticity} — can an observer accurately describe \emph{what} a feature represents? Together they form a comprehensive evaluation (see Fig.~\ref{fig:interface}, bottom): a feature can fire in a predictable location yet defy verbal description, or be easy to name yet hard to point to. The proposed framework has three properties that support a fair cross-model comparison: (i) it replaces the polysemantic neuron basis used in prior work with monosemantic directions recovered via sparse autoencoders~\citep{cunningham2023sparse, gao2024scaling}, so each tested unit corresponds to a single concept; (ii) it evaluates \emph{functionally analogous} features across models by anchoring feature selection to a shared set of input images (Fig.~\ref{fig:features}, top), rather than sampling features randomly from each model; and (iii) it uses calibrated scores to a common chance level across features and models, so scores are directly comparable. Furthermore, the evaluation is scaled to ${\sim}$6{,}000 features across six models at a fraction of the cost of previous protocols \citep{zimmermann2023}.

We apply the framework to six vision transformers—two supervised ViTs and four foundation models (DINOv2~\citep{oquab2023dinov2}, DINOv3~\citep{simeoni2025dinov3}, CLIP~\citep{radford2021learning}, SigLIP~\citep{zhai2023sigmoid})—and collect more than $15{,}000$ behavioral responses, of which the $13{,}400$ from the $377$ participants who passed our pre-specified quality checks enter the analyzes. The headline finding is sobering: every foundation model we tested produces features that are \emph{less} interpretable than those of the size-matched supervised baseline that came before, and downstream task performance does not predict the gap. Two representational properties do: feature \emph{locality} and \emph{coarse-grained} semantic alignment with human similarity judgments. DINOv3---whose training objective explicitly promotes local features---is the encouraging counterexample, suggesting that the gap is not architectural destiny.

Our contributions are: \textbf{(1)} a framework for measuring and comparing the human interpretability of vision models, with two complementary psychophysics protocols---\emph{localizability} and \emph{semanticity}; \textbf{(2)} a human study across six vision transformers showing that every foundation model we tested generally produces features that are \emph{less} interpretable than those of the size-matched supervised baselines that came before, yet this gap is not a consequence of their greater capability: downstream performance does not explain it; and \textbf{(3)} the identification of two representational properties that predict interpretability---\emph{locality} of feature activations and \emph{coarse-grained} semantic alignment with humans.

\section{Related Work}\label{sec:related}

\vspace{0mm}\paragraph{Sparse autoencoders and monosemanticity.}
Under the superposition hypothesis~\citep{elhage2022superposition}, networks encode more features than they have neurons by superimposing concepts onto shared activation channels, producing polysemantic neurons. Sparse autoencoders (SAEs) recover a dictionary of monosemantic directions and have emerged as the dominant remedy~\citep{cunningham2023sparse, gao2024scaling, templeton2024anthropicv2, fel2025archetypal}. We extract features via SAEs to avoid the polysemy that complicates neuron-level analysis.

\vspace{-1mm}\paragraph{Evaluation of model interpretability.}
\citet{bau2017network} introduced Network Dissection, to our knowledge the first framework to compare interpretability across models (architectures and training regimes). They score every convolutional unit from four CNNs against pixel-level masks from a set of human-labeled concepts (Broden), and models are ranked by the number of units that align with a concept above a given threshold. The approach is bounded by Broden's vocabulary, restricted to the neuron basis, and conflates concept alignment with interpretability---which, as Section~\ref{sec:strategies} shows, do not necessarily coincide. 

\citet{borowski2021} pioneered psychophysics-based evaluation of DNN interpretability: participants viewed maximally and minimally activating images for a given neuron, then identified which of two novel query images also strongly activated it. \citet{zimmermann2023} scaled this protocol across nine vision and vision-language architectures (80 features per model, 720 in total---a small fraction of the thousands a typical model encodes) and concluded that more capable models are not necessarily more interpretable. \citet{zimmermann2024measuring} later automated the evaluation by replacing human judgments with pairwise perceptual similarity. Their metric correlates with human results, but cross-model differences arise primarily from a tail of highly uninterpretable units linked to superposition. Applying a similar protocol to CNNs, \citet{colin2024choosing} showed that reliance on a neuron basis can mask cross-model differences because the impact of superposition is itself model-specific. Two methodological limitations, therefore, prevent fair cross-model comparison in this lineage: exclusive reliance on the polysemantic neuron basis and a forced-choice task whose process-of-elimination shortcuts produce model-specific chance levels. 
Our framework resolves both and extends model coverage to several foundation models that postdate prior studies.

% We build on this lineage with broader model coverage---several of the foundation models in our evaluation postdate prior studies---and a model-agnostic framework that resolves two limitations: exclusive reliance on the neuron basis, and a confound in the forced-choice task that produces model-specific chance levels and prevents fair cross-model comparison.

\section{Method}\label{sec:method}
\vspace{0mm}
\subsection{Current protocols are ill-suited for model interpretability.}

Prior protocols~\citep{borowski2021, zimmermann2023} frame the task as a contrastive forced-choice: participants view maximally and minimally activating images for a feature, then select which of two novel query images also strongly activates it (see Fig.~\ref{fig:normal} in Appendix~\ref{sec:control} for an example trial). While intuitive, this design has a fundamental confound: participants can identify the correct answer by understanding what the feature is not about---\emph{e.g.}, not a corn detector (Fig.~\ref{fig:normal})---rather than genuinely interpreting the feature itself.

To test whether this confounding effect is consequential, we performed a control experiment on two models from~\citet{zimmermann2023} (ResNet-50 and ViT-B32). We ran two versions of a given trial: the original version (Fig.~\ref{fig:normal}) and a variation where we replaced the maximally activating images with images ranking 25{,}000 out of 50{,}000 in each feature's activation distribution, thus carrying little to no information about the feature's visual content (Fig.~\ref{fig:control}). In this case, participants were expected to perform at chance ($50\%$). 
Participants nevertheless performed significantly above the $50\%$ chance level for both models. For ResNet-50, accuracy was $60.0\%$ across $2{,}480$ trials from 62 participants (binomial test, $95\%$ CI $[58.1\%,62.0\%]$, $p<.001$). For ViT-B/32, accuracy was $53.6\%$ across $1{,}840$ trials from 46 participants ($95\%$ CI $[51.3\%,55.9\%]$, $p=.001$). The protocol therefore produces significantly inflated and \emph{model-dependent baselines} even when the maximally activating images carry little information about the feature.

This model-specific baseline inflation makes raw accuracy scores incomparable across architectures.
\citet{zimmermann2023} reported ResNet-50 as more interpretable than ViT-B32 ($83\%$ vs.\ $80\%$). However, once each score is expressed relative to its model-specific baseline, the ranking \emph{reverses}: ViT-B32 is $27\%$ above its baseline while ResNet-50 is only $23\%$ above its own. Any protocol that does not account for this artifact risks drawing erroneous conclusions.

\vspace{0mm}
\subsection{Framing interpretability as identification.}
\vspace{0mm}\paragraph{Overview.}
We address the limitations described above with a framework comprising of two complementary protocols that operationalize interpretability: \emph{localizability}, \emph{i.e.}, can a participant predict \emph{where} a feature fires on a novel image?; and \emph{semanticity}, \emph{i.e.}, can a participant accurately describe \emph{what} a feature represents? 
Both protocols share the same feature extraction pipeline (Fig.~\ref{fig:methodo} in Appendix~\ref{app:methodo}) and the same visual explanations presented to participants (Fig. \ref{fig:interface}). They differ only in the task and scoring function. Because there is no distractor to exploit, participants must rely on genuine understanding of the feature's visual content to perform. Localizability scores are mapped to a common chance-anchored scale (Section~\ref{sec:scoring}), enabling direct comparison across features, models and protocols. Semanticity requires no chance-level correction: images and descriptions vary across models, but they are scored in a shared embedding space, making cross-model comparisons fair.

By evaluating the interpretability of features, we aim to derive a more general score representative of model interpretability. Whereas~\citet{zimmermann2023} studied features across multiple layers, our focus on foundation models leads us, moving forward, to examine only the last latent representation of each model, \emph{i.e.}, the output of the penultimate layer for supervised models, the output of the last layer for DINO models, and the output of the vision encoder for VLMs. 

\vspace{0mm}\paragraph{Feature extraction.} \label{sec:method_details}
For each model, we extract features from its vision encoder. Because individual neurons in vision models are often polysemantic~\citep{elhage2022superposition}, we use sparse autoencoders (SAEs) to recover a dictionary of monosemantic directions. TopK SAEs~\cite{gao2024scaling} (K=38) are trained on activations from the ImageNet training split (${\approx}1.28$M images), using all patch tokens per image (256 for patch size 14 in DINOv2, 196 for patch size 16 in all other models). Each SAE uses an expansion factor of $\times 10$, yielding $3{,}840$ features for ViT-S and $7{,}680$ features from all the other models. Reconstruction quality, dead-feature rates, and stability across independent SAE seeds are reported in Appendix~\ref{app:sae_quality}.

\vspace{0mm}\paragraph{Feature relevance and linear probing.}
For each model, we train a linear classification head on ImageNet~\cite{imagenet_cvpr09} on top of the frozen backbone, from which we derive (i) per-feature importance scores that quantify each feature's contribution to the classifier's predictions, computed as Gradient input~\cite{shrikumar2017learning}, shown to be particularly faithful in latent space~\cite{fel2023holistic}; and (ii) a downstream accuracy baseline used in Section~\ref{sec:im1k}.

\vspace{0mm}\paragraph{Feature selection.}
Given that current computer vision models are expected to encode thousands of features, we aim to scale the number of features evaluated. Building on \citet{zimmermann2023}, we first ran a pilot with $80$ features and $10$ trials per feature to obtain reliable per-feature interpretability estimates and a model interpretability score. We then evaluated the depth-vs-breadth tradeoff: we found that the number of features evaluated is more important than the number of repetitions per feature to obtain stable model-level estimates (see Appendix~\ref{app:morefeatures}). We therefore adopted this design in the main experiment. 

To obtain a representative subset of features, we aim to select features that are both diverse and relevant to the model’s decisions. To this end, we randomly sample one image for each of the 559 superordinate classes underlying the 1,000 ImageNet labels, yielding a fixed set of 559 images. For each image, we select the three visual features that rank as the most important for the model’s prediction on that image.  During selection, we exclude features that activate on more than $10\%$ of patch tokens, following~\citet{sun2025dense}---in practice, this removes about 7-90 candidate features per model. This procedure yields a total of $559 \times 3 = 1{,}677$ trials per model, but because the same feature can be selected for multiple images, this results in 500--1,300 unique features evaluated per model (see Appendix~\ref{sec:reuse}).

\vspace{0mm}\paragraph{Feature explanation.} Both protocols present participants with the same visual explanation of a feature (Fig.~\ref{fig:interface}- top): (a) a synthetic feature visualization synthesized via MACO~\citep{fel2023maco} that maximally activates the feature (left panel in Fig.~\ref{fig:interface}); (b) nine highly activating natural images (middle panel in Fig.~\ref{fig:interface}); and (c) their corresponding RISE heatmaps~\citep{petsiuk2018rise} indicating where the feature fires on each image (right panel in Fig.~\ref{fig:interface}). Participants use these three sources of information to form an understanding of the feature before solving the task.

\vspace{0mm}\paragraph{Localizability (\emph{where}).}\label{sec:scoring}

Each \emph{localizability} trial additionally presents (d) a novel query image (bottom, left in Fig.~\ref{fig:interface}) and  participants are asked to click on the region of the image where they believe the feature appears. Each click is scored against the RISE heatmap of the query image (smoothed with a Gaussian filter). Since an attribution map is only a proxy ground truth, we test whether the resulting comparison depends on this choice by rescoring the original human clicks with HSIC~\citep{novello2022making}, an independent black-box attribution method. The broad model ordering is preserved, indicating that the results presented below are not specific to RISE; full results are reported in Appendix~\ref{app:hsic_robustness}.

Let $v$ be the heatmap value at the clicked location and $p = \widehat{F}(v)$ its empirical cumulative distribution function (ECDF) over the heatmap pixels. Because heatmap distributions vary across features (\emph{e.g.}, sparse vs.\ diffuse activations), raw percentiles are not directly comparable across features or models.
We therefore introduce a \emph{chance-anchored normalization} that pins chance to $s = 0.5$. Writing $p_{\mu} = \widehat{F}(\mu)$ for the percentile rank of the mean activation $\mu$, we map $p_{\mu}$ to $0.5$ and the extremes to $0$ and $1$:

\begin{equation*}
    s(v)
=
\begin{cases}
0.5 - 0.5 \cdot \dfrac{p_{\mu} - p}{p_{\mu}}, & \text{if } p < p_{\mu}, \\[10pt]
0.5 + 0.5 \cdot \dfrac{p - p_{\mu}}{1 - p_{\mu}}, & \text{otherwise.}
\end{cases}
\end{equation*}

By pinning chance ($s = 0.5$) to the mean activation, the score carries the same interpretation across features and models — a click above the mean scores above chance, below the mean scores below---regardless of whether the feature activates focally or diffusely. We refer to this score as the \emph{localizability} score.

\vspace{0mm}\paragraph{Semanticity (\emph{what}).} 

Each \emph{semanticity} trial asks participants to (d) write a short free-text description of the feature (bottom right in Fig.~\ref{fig:interface}) and rate their confidence on a 5-point Likert scale. We evaluate two complementary properties. Description \emph{plausibility} measures whether the proposed interpretation is compatible with the feature's visual evidence, while descriptions \emph{agreement} measures whether independent participants converge on a similar interpretation. \emph{Confidence} helps distinguish ambiguity, reflected by low agreement and low confidence, from confident misinterpretation, reflected by low plausibility despite high confidence.

To measure description \emph{plausibility}, we extract a $96\times96$ crop around the peak RISE activation from each of 
the nine most-activating images  and encode these crops and the participant's description using CLIP ViT-B/16~\citep{radford2021learning}. The score is the mean cosine similarity between the text embedding and the nine images embeddings. We validate that the results are stable across six alternative vision-language encoders ($\rho=.74$--$.94$, all $p<.001$) and when using full images rather than crops; see Appendix~\ref{sec:semanticity_details}.

To measure descriptions \emph{agreement}, we embed all descriptions of features with at least two responses using a popular sentence-encoder, MPNet~\citep{song2020mpnet}, and compute their mean pairwise cosine similarity. High agreement indicates a clear, shared interpretation, whereas low agreement suggests that the feature is ambiguous..

Because agreement requires multiple descriptions per feature, we evaluate a representative subset. For each model, we divide features into localizability deciles and sample uniformly within each bin, yielding approximately 30 features per decile and 300 per model, leading to approximately  900 \emph{semanticity} trials per model. Due to the CLIP modality gap, \emph{plausibility} scores range from $0.13$ to $0.35$, with a chance baseline of $0.19$, defined as the average similarity between image and text embeddings drawn from different features.

\vspace{0mm}\paragraph{Participants.}
Across both experiments, we recruited about $440$ participants via Prolific\footnote{\url{www.prolific.com}}. All were native English speakers without reported visual impairments and completed the study on a laptop or desktop. Each provided informed consent and received \$2.95 (${\sim}\$16$/hr; 10--13 minutes). The protocol was approved by the IRB of an author-affiliated institution. We analyze about $13,400$ behavioral responses from the $377$ participants who (i) passed at least 4 of 6 practice trials, (ii) answered all 4 attentiveness catch trials correctly, and (iii) completed the experiment within 3 SDs of the mean duration. We define one \emph{behavioral response} as one participant's answer to one trial: a click in the localizability experiment or a free-text description, together with its confidence rating, in the semanticity experiment.

\section{Results}

We evaluated six vision transformers: two supervised baselines (\texttt{ViT-S/16} and \texttt{ViT-B/16}) and four foundation models---\texttt{DINOv2 ViT-B/14}, \texttt{DINOv3 ViT-B/16}, \texttt{CLIP ViT-B/16}, and \texttt{SigLIP ViT-B/16}.

\begin{table}[ht!]
\centering
\caption{Localizability, Semanticity (plausibility and agreement) and confidence scores across six vision transformers. The size-matched supervised model (ViT-B) consistently outperforms foundation models on both dimensions. Best result per row is in bold, second best is underlined.}
\label{tab:scores}
\vspace{1mm}
\small
\begin{tabular}{lc!{\color{gray!25}\vrule width 0.4pt}ccccc}
\toprule
 & \multicolumn{2}{c}{\textbf{Supervised}} & \multicolumn{4}{c}{\textbf{Foundation}} \\
\cmidrule(lr){2-3}\cmidrule(lr){4-7}
 & ViT-S & ViT-B & DINOv2 & DINOv3 & CLIP & SigLIP \\
\midrule
Localizability ($\uparrow$)
& \underline{80.3}
& \textbf{86}
& 71
& 79.6
& 79.7
& 71.4 \\

Semanticity (plausibility) ($\uparrow$)
& \textbf{0.274}
& \underline{0.273}
& 0.259
& 0.260
& 0.266
& 0.253 \\

Semanticity (agreement) ($\uparrow$)
& \underline{0.493}
& \textbf{0.519}
& 0.433
& 0.458
& 0.440
& 0.443 \\

Semanticity confidence ($\uparrow$)
& 3.43
& \underline{3.61}
& \textbf{3.68}
& 3.41
& 3.51
& 3.38 \\
\bottomrule
\end{tabular}
\end{table}

\vspace{0mm}\paragraph{Foundation models are less interpretable than supervised ones.}\label{sec:headline}

We measure two dimensions of interpretability: \emph{localizability} and \emph{semanticity}, the latter quantified through description \emph{plausibility} ($P$) and \emph{agreement} ($A$)\footnote{We report the median per model because the scores are not normally distributed; see Appendix~\ref{app:skew}.}, as summarized in Table~\ref{tab:scores}. Scores differ significantly across models ($L$: Kruskal--Wallis, $H(5)=118.40$, $p<.001$; $P$: $H(5)=295.36$, $p<.001$; $A$: $H(5)=30.3$, $p<.001$). As seen in the table, foundation models are consistently less interpretable than their size-matched supervised counterparts (Dunn's test and Tukey HSD's test, $p<.001$ vs.\ ViT-B), regardless of whether they were trained with language supervision (CLIP, SigLIP) or pure self-supervised objectives (DINOv2, DINOv3). Moreover, while we expect \emph{localizability} and \emph{semanticity} to quantify two different dimensions of interpretability, in practice we find that they are highly correlated ($L$-$P$: $r=0.84$, $p=0.036$; $L$-$A$: $r=0.82$, $p=0.046$; $P$-$A$: $r=0.80$, $p=0.059$). Qualitative examples spanning the four quadrants of \emph{localizability} and \emph{plausibility} are shown in Appendix Fig.~\ref{fig:quadrant_loc_name}.

\vspace{0mm}\paragraph{Models can feel more interpretable than they are.}\label{sec:confidence}

Average confidence is fairly homogeneous across models ($3.38-3.68$, Table~\ref{tab:scores}, bottom row), but its alignment with interpretability is not ($P$, $\rho=0.34$; $A$, $\rho=0.30$). While DINOv2 elicits the highest confidence and ranks near the bottom on interpretability (last on \emph{localizability} and \emph{agreement}, second-to-last on \emph{plausibility}), SigLIP performs poorly on all fronts. Based on these results, a model can feel more interpretable than it is---see Appendix Fig.~\ref{fig:quadrant_conf_name} for examples of features that are confidently but inaccurately named (and vice versa), a caution worth keeping in mind given that DINOv2 features have been treated as a reference for interpretable representations~\citep{dreyer2024pure}.

% \vspace{2mm}

% Given the strong correlation between \emph{plausibility} and \emph{agreement}, for the remainder of the analyses, we focus only on the former and refer to it more generally as \emph{semanticity}.

\subsection{Alignment with performance} 
\begin{figure}[t]
  \centering
\includegraphics[width=\textwidth]{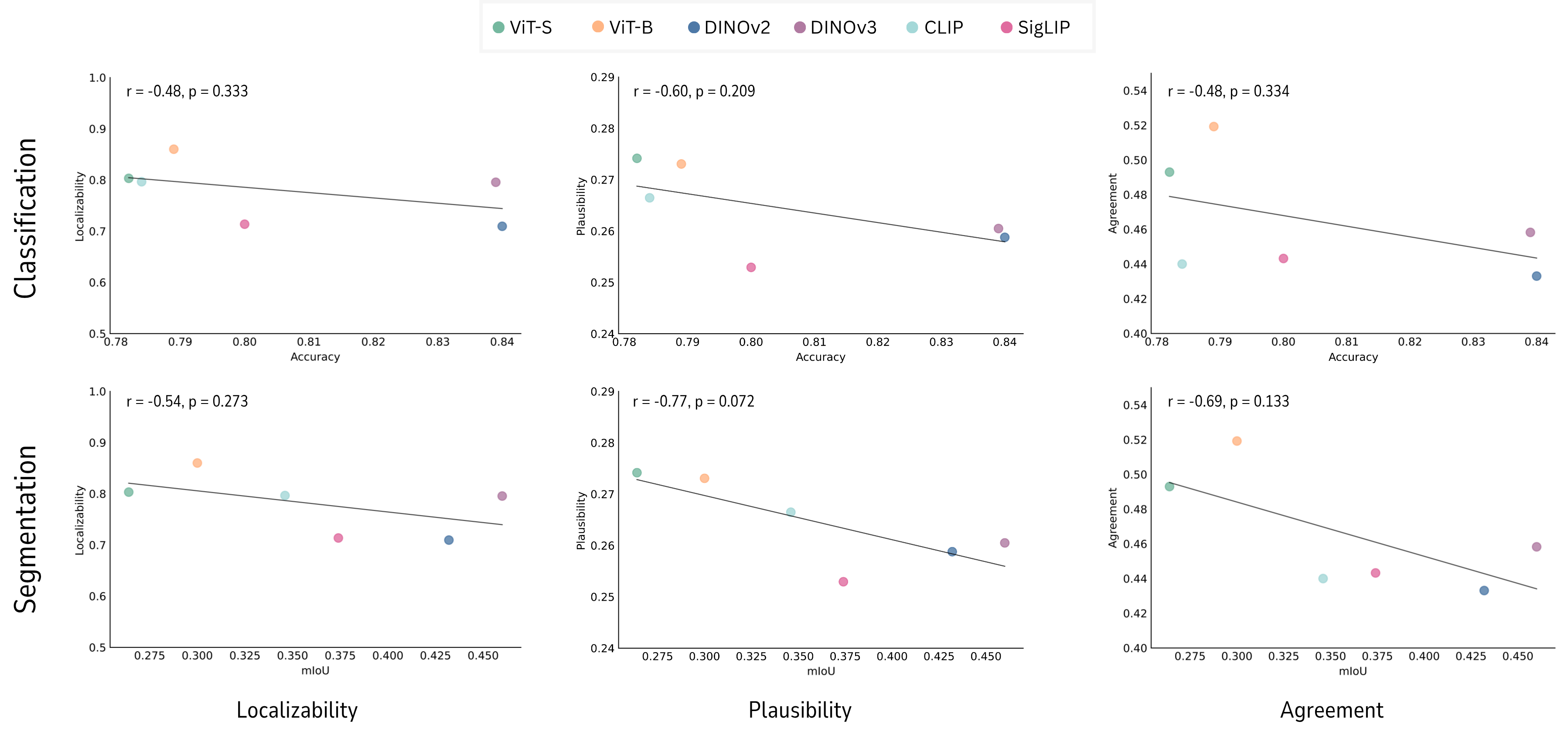}
  \caption{\textbf{Interpretability is uncorrelated with downstream task performance.} Each panel plots an interpretability score against a capability benchmark across the six models. Top row: Classification (ImageNet-1k top-1 accuracy~\cite{imagenet_cvpr09}); bottom row: Segmentation (ADE20K semantic segmentation~\cite{zhou2017ade20k}). Columns: \emph{Localizability}, \emph{Plausibility}, and \emph{Agreement}. Spearman $\rho$ and $p$-values are shown in each panel. Correlations are non-significant for all three interpretability measures across both benchmarks, suggesting that interpretability and downstream task performance capture largely orthogonal properties of a visual representation.}\label{fig:performance}
\end{figure}

A natural question is whether interpretability and capability are linked: models with richer, more semantically structured representations might be both more capable on downstream tasks and more legible to humans. We measured the Spearman correlation between each model's interpretability scores and its performance on three vision tasks: ImageNet-1k classification, semantic segmentation, and perceptual grouping. We report the results for the first two in Fig.~\ref{fig:performance}.

\vspace{0mm}\paragraph{ImageNet-1k classification.}\label{sec:im1k}
Foundation models lead on ImageNet top-1 accuracy but trail on interpretability. Across the six models, neither \emph{localizability} ($\rho = -0.48$, $p = 0.333$) nor \emph{semanticity} (P: $\rho = -0.60$, $p = 0.209$; A: $\rho = -0.48$, $p = 0.334$) correlate significantly with classification. DINOv2 and SigLIP match CLIP on accuracy yet score roughly 8 points lower in \emph{localizability}.

\vspace{0mm}\paragraph{Semantic segmentation.}
Segmentation performance (mIoU) on ADE20K\cite{zhou2017ade20k}---using a linear probe\footnote{We follow the DINOv3 methodology but force all models to an input size of 224 $\times$ 224, since supervised ViTs from \texttt{timm} cannot take flexible input sizes. The same pipeline at 512 $\times$ 512 reproduces DINOv3 results.}---does not correlate significantly with either \emph{localizability} ($\rho = -0.54$, $p = 0.273$) or with \emph{semanticity} (P: $\rho = -0.77$, $p = 0.072$; A: $\rho = -0.69$, $p = 0.133$).

\vspace{0mm}\paragraph{Perceptual grouping.}
On a perceptual grouping benchmark\footnote{Perceptual grouping refers to the object-instance binding benchmark of~\citet{li2026binding}: the ability to determine which patches or regions belong to the same object instance, measured through the benchmark's binding-accuracy protocol.}~\cite{li2026binding}---a more constrained form of segmentation that probes the models' ability to bind individual object instances---we again find no significant correlation with either \emph{localizability} ($\rho = 0.12$, $p = 0.820$) nor \emph{semanticity} (P: $\rho = 0.09$, $p = 0.867$; A: $\rho = -0.12$, $p = 0.817$).

\textit{Across all three benchmarks, task performance does not predict interpretability}. This dissociation suggests that the representational properties that make a model useful for downstream tasks are largely orthogonal to those that make its features interpretable to humans.

\vspace{0mm}\paragraph{Locality of the representation.}\label{sec:sparsity}
Beyond task performance, we ask whether \emph{locality}, \emph{i.e.}, the degree to which a feature's activations concentrate on confined image regions, predicts interpretability: a focal activation supplies a concrete spatial anchor (an object part, a texture), while a diffuse one leaves the participant to guess what the activating regions share. We quantify locality with the Hoyer metric applied to each feature's RISE heatmaps~\citep{hoyer2004non}:
\begin{equation*}
   H(\mathbf{x}) = \frac{(\sqrt{n} - \|\mathbf{x}\|_1/\|\mathbf{x}\|_2)}{(\sqrt{n} - 1)}
\end{equation*}

with $n$ the number of pixels, and where $H = 0$ corresponds to uniform activation across the image and $H = 1$ to all mass on a single pixel. We average $H$ across each feature's 9 most activating images and then across the model´s features. In contrast to task performance, locality correlates strongly with \emph{localizability} ($\rho = 0.91$, $p = 0.011$;  Fig.~\ref{fig:local_coarse}, top left) and \emph{semanticity} (P: $\rho = 0.99$, $p<.001$; Fig.~\ref{fig:local_coarse}, top middle; A: $\rho = 0.82$, $p=0.048$; Fig.~\ref{fig:local_coarse}, top right). See Appendix~\ref{app:kolmogorov} for additional results.

\subsection{Alignment with human perception}\label{sec:alignment}

Next, we consider human alignment, \emph{i.e.}, the degree to which a model aligns with some properties of human perception, as alignment has been shown to improve representation quality over a wide range of downstream tasks~\cite{sucholutsky2023alignment, muttenthaler2023improving, sundaram2024dreamsimv2, muttenthaler2025nature}. Because not all measures of alignment capture the same properties of human perception~\cite{ahlert2024aligned}, we study 3 kinds of alignment: alignment with human (a) visual strategies; (b) coarse-grained similarity judgments (semantic similarity); and (c) fine-grained similarity judgments (lower-level visual similarity).

\vspace{0mm}\paragraph{Alignment with human visual strategies.}
Strategy alignment---the overlap between model-driven and human-attended image regions ~\citep{linsley2018learning,fel2022harmonizing}---does not correlate significantly with \emph{localizability} ($\rho = 0.14$, $p = 0.798$) or \emph{semanticity} (P: $\rho = 0.21$, $p = 0.696$; A: $\rho = 0.30$, $p = 0.564$). 

\vspace{0mm}\paragraph{Alignment with human similarity judgments.} \label{sec:strategies}
Human similarity judgments operate at multiple levels of abstraction~\citep{muttenthaler2025nature}.
\emph{Fine-grained} similarity captures within-category distinctions (\emph{e.g.}, how similar two butterfly species are to each other), while \emph{coarse-grained} similarity captures broad semantic structure across categories (\emph{e.g.}, how similar a buffalo is to a spider relative to grass).
We measure alignment at the different levels of similarity using three datasets: THINGS similarity~\citep{hebart2020revealing, hebart2023things}, Levels~\cite{muttenthaler2025nature}, and NIGHTS~\citep{fu2023dreamsim, sundaram2024dreamsimv2}.

Alignment with \emph{fine-grained} similarity does not correlate with neither \emph{localizability} (NIGHTS: $\rho = -0.41$, $p = 0.418$; Levels: $\rho = 0.20$, $p = 0.705$), \emph{plausibility} (NIGHTS: $\rho = -0.47$, $p = 0.347$; Levels:$\rho = 0.26$, $p = 0.617$) or \emph{agreement} (NIGHTS: $\rho = -0.33$, $p = 0.523$; Levels:$\rho = -0.06$, $p = 0.915$).

\begin{figure}[ht!]
  \centering
\includegraphics[width=\textwidth]{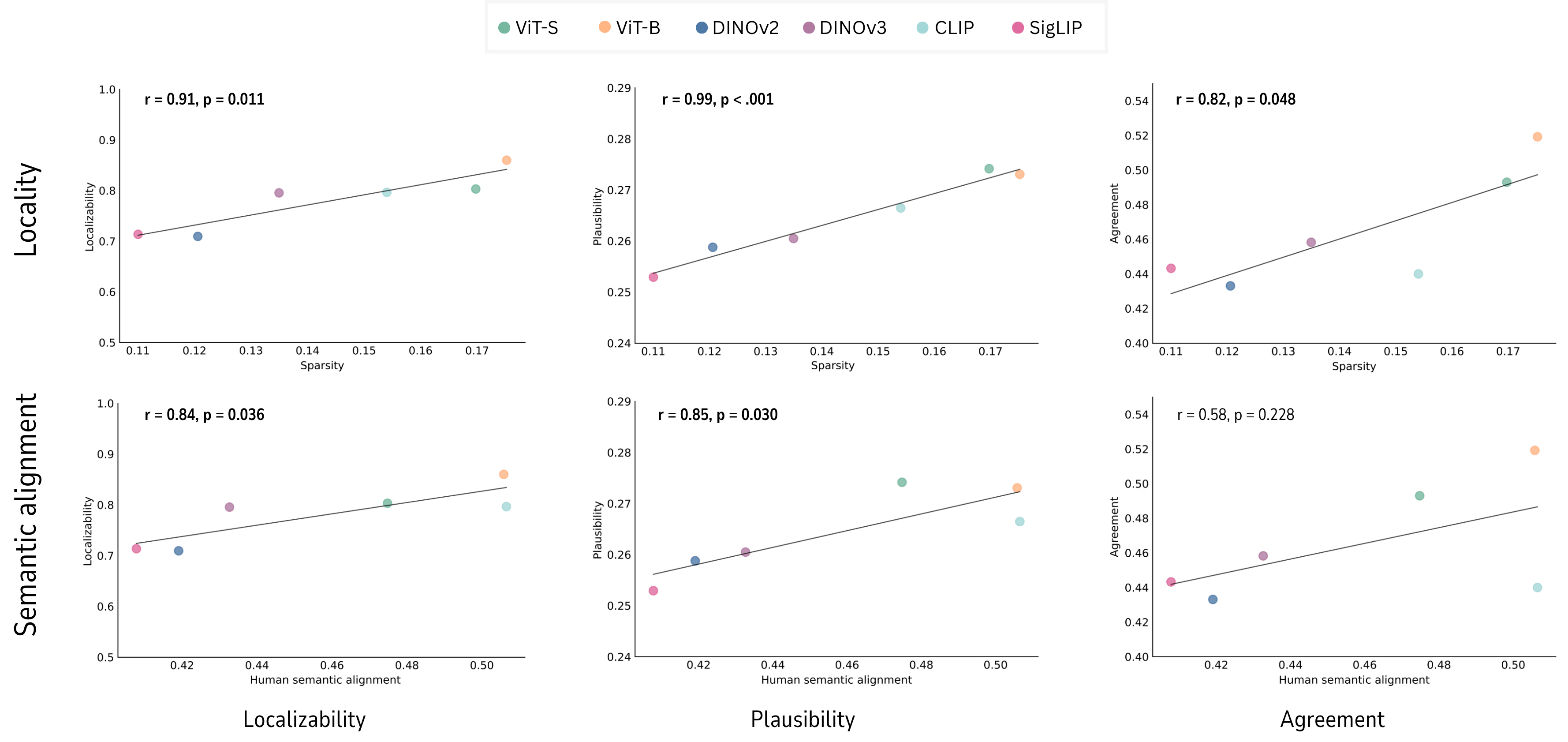}
\caption{\textbf{Locality and coarse-grained semantic alignment track interpretability.} Each panel plots an interpretability score against a representational property across the six models. Top row: Locality ((mean Hoyer sparsity over feature heatmaps); bottom row: Human semantic alignment (coarse-grained alignment with human similarity judgments on THINGS~\cite{hebart2020revealing}). Columns: \emph{Localizability}, \emph{Plausibility}, and \emph{Agreement}. Spearman $\rho$ and $p$-values per panel. Locality is the strongest predictor of interpretability, consistent with focal activations providing a clearer visual anchor for understanding a feature. Coarse-grained alignment also tracks interpretability, particularly for \emph{localizability} and \emph{plausibility}, suggesting that interpretable representations are not only local but also organized in ways that better reflect human categorical similarity judgments.}
  \label{fig:local_coarse}
\end{figure}

In contrast, alignment with \emph{coarse-grained} similarity is generally lower and shows a strong correlation with interpretability (Fig. \ref{fig:local_coarse}, bottom), which is significant for the THINGS similarity dataset for both \emph{localizability} and \emph{plausibility}: \emph{localizability} (THINGS: $\rho = 0.84$, $p = 0.036$, Levels:$\rho = 0.71$, $p = 0.110$), \emph{plausibility} (THINGS: $\rho = 0.85$, $p = 0.030$; Levels $\rho = 0.6$, $p = 0.205$), and \emph{agreement} (THINGS: $\rho = 0.85$, $p = 0.03$; Levels $\rho = 0.6$, $p = 0.21$). See Appendix~\ref{app:levels-class} for additional results.

This dissociation between granularity levels suggests that what matters for interpretability is not perceptual fidelity to fine visual detail, but rather the degree to which a model organizes its features around the same high-level semantic categories that structure human perception.
A feature that groups concepts according to what humans consider broadly similar is more likely to be a feature that humans can understand.

\section{Discussion}

\vspace{0mm}
\paragraph{Two protocols, three measures, one interpretability.}
The two protocols differ in nearly every respect: one asks participants to point, the other to write; one is scored against a spatial heatmap, the other against a language or vision--language embedding. That they yield concordant rankings and surface the same predictors is unlikely to be a coincidence---we appear to be measuring a stable property of the representations rather than an artifact of any one task, and either protocol may suffice on its own in future work.

\vspace{0mm}
\paragraph{Foundation models are less interpretable than their size-matched supervised counterparts.}
The gap appears in both vision-only SSL models (DINOv2, DINOv3) and vision-language models (CLIP, SigLIP), so neither pretraining signal is the cause. The common factor is foundation-style pretraining itself. One reading is that these models reach---and on many benchmarks exceed---human performance in part by developing \emph{superhuman} features: representations effective for the task but not the kind a human observer can readily make sense of. The very generality that makes foundation models capable may also push their representations away from a human-interpretable basis. This gap is easy to miss without direct evaluation---DINOv2, often treated as a reference for interpretable representations~\citep{dreyer2024pure}, elicits the highest rater confidence yet ranks among the least interpretable models on \emph{localizability} and \emph{semanticity}.

\vspace{0mm}
\paragraph{Interpretability is orthogonal to task performance.}
Across every benchmark we examined, capability and interpretability are uncorrelated. The relationship is neither a tradeoff nor a free lunch: how well a model classifies, segments, or groups says little about whether a human can interpret its features. The leaderboards by which the field currently judges progress are silent on interpretability, a property independently desirable in high-stakes settings (\emph{e.g.}, clinical decision support or autonomous driving) where these same backbones are increasingly considered. Interpretability has to be designed for explicitly; it will not arrive as a byproduct of capability, and prior evidence suggests that scale alone does not guarantee it either~\cite{zimmermann2023}.

\vspace{0mm}
\paragraph{Locality is the proximate signature of an interpretable representation.}
The locality of feature activations is the strongest predictor of interpretability we identify. Foundation models tend to develop diffuse features that fire across larger, less focused regions, blending local content with broader scene context. This may help on downstream tasks but leaves a human observer without a clear visual anchor. DINOv3 is the encouraging exception: one of the most capable models in our set and also among the most interpretable foundation models, a pattern consistent with its training objective explicitly promotes local features. This finding suggests that locality-aware objectives can partially closes the interpretability gap, though whether they do so without capability cost requires additional direct experimental verification.

\vspace{0mm}
\paragraph{Coarse-grained semantic alignment, not perceptual fidelity, predicts interpretability.}
Of the alignment measures we considered, only coarse-grained alignment with human similarity judgments tracks interpretability. A model can attend to the same image regions humans do, yet still encode what it sees there in features humans cannot make sense of. Fine-grained perceptual alignment has been found to be uniformly high and largely shared across models~\citep{groger2026aristotelian}. Coarse-grained alignment, on the other hand, is consistently lower, leaving real room for improvement. Interpretability appears driven less by perceptual fidelity than by how a representation organizes the visual world at the categorical level. Training signals built from human odd-one-out judgments~\citep{hebart2020revealing} seem like a promising direction to close part of the gap.

\section{Conclusion}
We introduced a framework for measuring and comparing the human interpretability of vision models, built around two complementary psychophysics protocols: \emph{localizability} (where a feature fires) and \emph{semanticity} (what it represents). Across six vision transformers and more than $15{,}000$ behavioral responses, foundation models, whether trained with language supervision or self-supervised objectives, were consistently \emph{less} interpretable than their size-matched supervised counterparts---a surprising result for models that have surpassed those same baselines on most other axes. The gap is not a capability tradeoff: interpretability is uncorrelated with downstream task performance on every benchmark we examined, and the two structurally different protocols agree on this conclusion and on the predictors that drive it. Two properties track interpretability: the \emph{locality} of a feature's activations and \emph{coarse-grained} semantic alignment with human similarity judgments. Together, they point to a concrete recipe for building more interpretable vision models---training objectives that promote local feature activations and align representations with the coarse semantic structure of human perception~\citep{muttenthaler2025nature}---and DINOv3, whose objective explicitly promotes locality, suggests that capability and interpretability need not trade off when locality is an explicit training objective.

\section{Acknowledgments}
The authors thank Michael Lepori and Grégoire Dhimoila for helpful discussions during the course of this project.
T.S. acknowledges support from the Office of Naval Research (N00014-24-1-2026) and the National Science Foundation (IIS-2402875). J.C. and N.O. acknowledge support from the Regional Government of Valencia, Spain (Conselleria de Industria, Turismo, Innovación y Comercio, Dirección General de Innovación), the European Union's Horizon Europe research and innovation program (ELIAS, grant agreement 101120237), and Intel Corporation; J.C. further acknowledges the Banco Sabadell Foundation. \\ \\
%% ---------------------------------------------------------------------------
%% CAMERA-READY FUNDING ACKNOWLEDGMENT — keep commented while under review.
%% Uncomment the block below at camera-ready (added 2026-08-13, T. Serre).
%%
% This work was supported by the Office of Naval Research (N00014-24-1-2026),
% the ONR REPRISM MURI (N00014-24-1-2603), and the National Science Foundation
% (IIS-2402875). This research was conducted using computational resources and
% services at the Center for Computation and Visualization, Brown University,
% supported by the Office of the Director of the National Institutes of Health
% under award number S10OD036341 (``High-Performance Compute Cluster for Brain
% Science''). The content is solely the responsibility of the authors and does
% not necessarily represent the official views of the National Institutes of
% Health. Co-authors affiliated with ANITI acknowledge the Artificial and
% Natural Intelligence Toulouse Institute (ANR-19-PI3A-0004).
%% ---------------------------------------------------------------------------

\bibliography{references}
\bibliographystyle{unsrtnat}

\newpage
\appendix
\setcounter{figure}{0}
\renewcommand{\thefigure}{A.\arabic{figure}}
\setcounter{table}{0}
\renewcommand{\thetable}{A.\arabic{table}}

\section{Limitations and broader impact.}

\paragraph{Limitations.} We acknowledge several limitations. First, our model-level analyses span only six architectures, which limits statistical power and makes it difficult to disentangle correlated properties of representations (\emph{e.g.}, sparsity and coarse-grained alignment may co-vary across architectures). Second, almost all models are evaluated at ViT-B scale. This controls for scale across architectures, but whether the interpretability gap between supervised and foundation models persists or reverses at larger scales remains an open question. Third, we focus on a specific set of representational properties: downstream task performance, locality, and alignment with human similarity judgments. Other factors, such as training data or robustness, may also shape interpretability.
Finally, our \emph{semanticity} protocol relies on external vision-language models to score the plausibility of human descriptions. This is imperfect: using an external model introduces its own biases and cannot provide a direct ground-truth measure of whether a description captures the feature. However, directly testing free-form descriptions against the evaluated vision foundation models is not currently possible, since these models do not provide a reliable way to judge whether a natural-language description matches one of their internal visual features. We therefore use external image--text similarity as a practical proxy, keeping the scorer fixed across all evaluated models so that its biases are held constant in the comparisons.

\paragraph{Broader impact.}
This work contributes to human-centered evaluation of model interpretability. By testing whether people can localize and describe decision-relevant features, our protocol adds to a broader effort to assess whether vision models are transparent enough for settings where human oversight matters, including critical applications. However, interpretability evaluations should not be taken as evidence of safety or reliability on their own. A model whose features appear more interpretable may still fail under distribution shift or behave unpredictably in downstream applications. Our results should therefore be viewed as one component of a broader evaluation ecosystem, complementing robustness, fairness, calibration, and task-specific validation.

\newpage
\section{Additional Related Work}
\vspace{0mm}\paragraph{Representational alignment with human perception.}
A growing body of work investigates how well DNN representations mirror human perceptual judgments~\citep{sucholutsky2023getting}, with parallel evidence that aligning model feature attributions with human importance maps improves object recognition~\citep{fel2022harmonizing}. Alignment with human similarity judgments has been shown to improve transfer~\citep{sucholutsky2023alignment, muttenthaler2023improving}, though benefits are task-dependent: alignment helps retrieval but may hurt discriminative classification~\citep{sundaram2024dreamsimv2}. Crucially, \citet{muttenthaler2025nature} showed that this trade-off depends on granularity: coarse-grained and fine-grained alignment affect downstream tasks differently. While prior work treats alignment as a driver of task performance, we treat it as a potential correlate of \emph{feature interpretability}---a connection that, to our knowledge, has not been investigated.

\vspace{0mm}\paragraph{Automated feature scoring with vision-language models.}
\citet{hernandez2022natural} proposed labeling visual neurons with natural-language descriptions and measuring label quality via retrieval accuracy. \citet{bills2023language} extended this idea to language model neurons, using GPT-4 to generate explanations scored by how well they predict held-out activations. CLIP-Dissect~\citep{oikarinen2023clipdissect} matched each neuron to the most CLIP-similar concept from a large vocabulary, enabling scalable, open-vocabulary labeling without human annotation. Our \emph{semanticity} score builds on this spirit---participants describe a feature in free text and we measure CLIP cosine similarity between their description and feature's preferred images---while retaining the validity guarantee of human behavioral data.

\newpage
\section{Details about the control experiment}\label{sec:control}

\begin{figure}[ht!]
  \centering
\includegraphics[width=\textwidth]{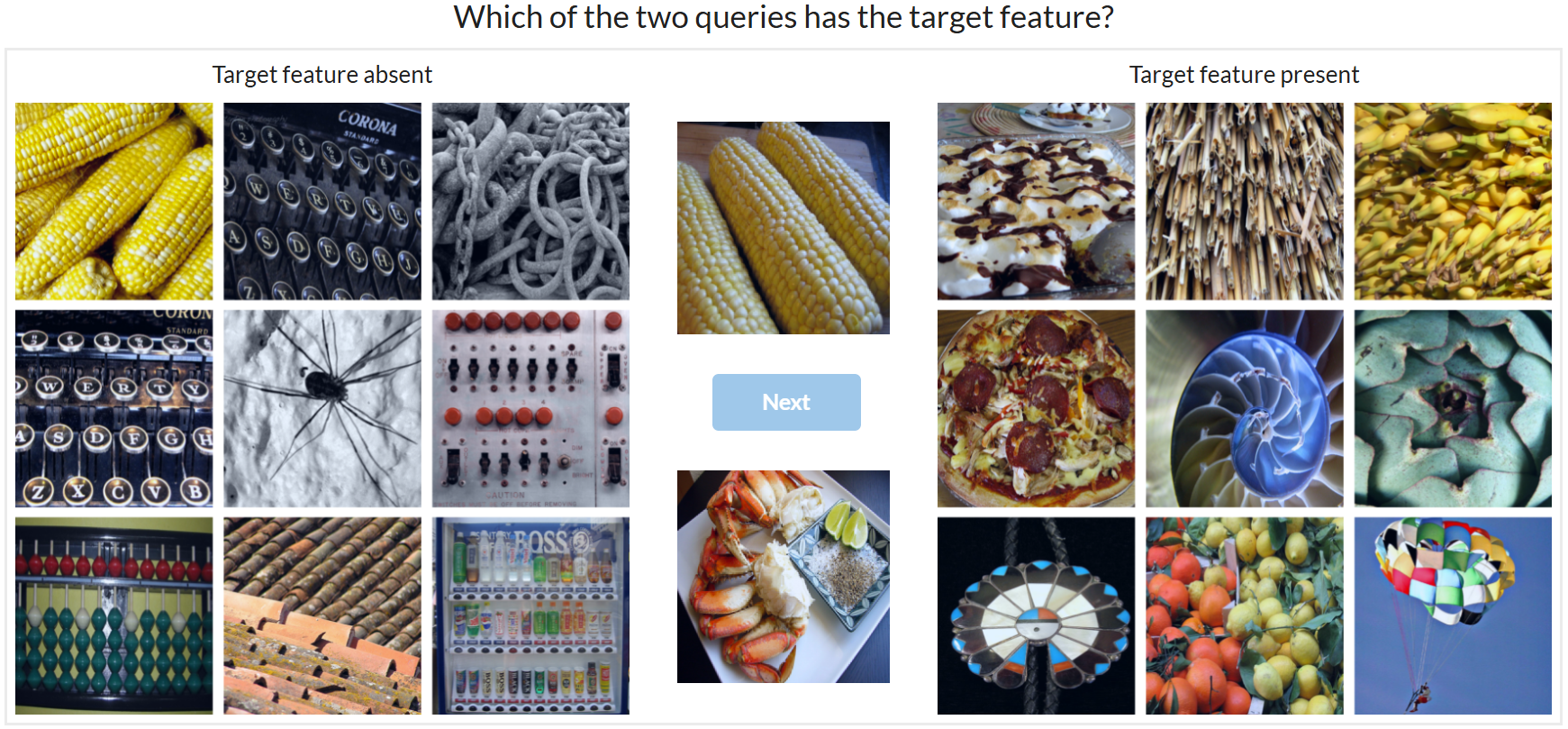}
  \caption{Example of a trial from~\citet{zimmermann2023}. In this protocol, the task is to pick the query image that matches the reference images illustrating the feature (left panel), by leveraging reference images that both highly activate the feature (right panel) and images that minimally activate it (left panel). In this example, the task can be trivially solved by relying on semantic grouping. By observation of the minimally activating stimuli (left panel), it is easy to conclude that the neuron of interest is not a corn detector, yet, it is hard to articulate what visual pattern is captured by the neuron (images in the right panel).}\label{fig:normal}
\end{figure}

\begin{figure}[ht!]
  \centering
\includegraphics[width=\textwidth]{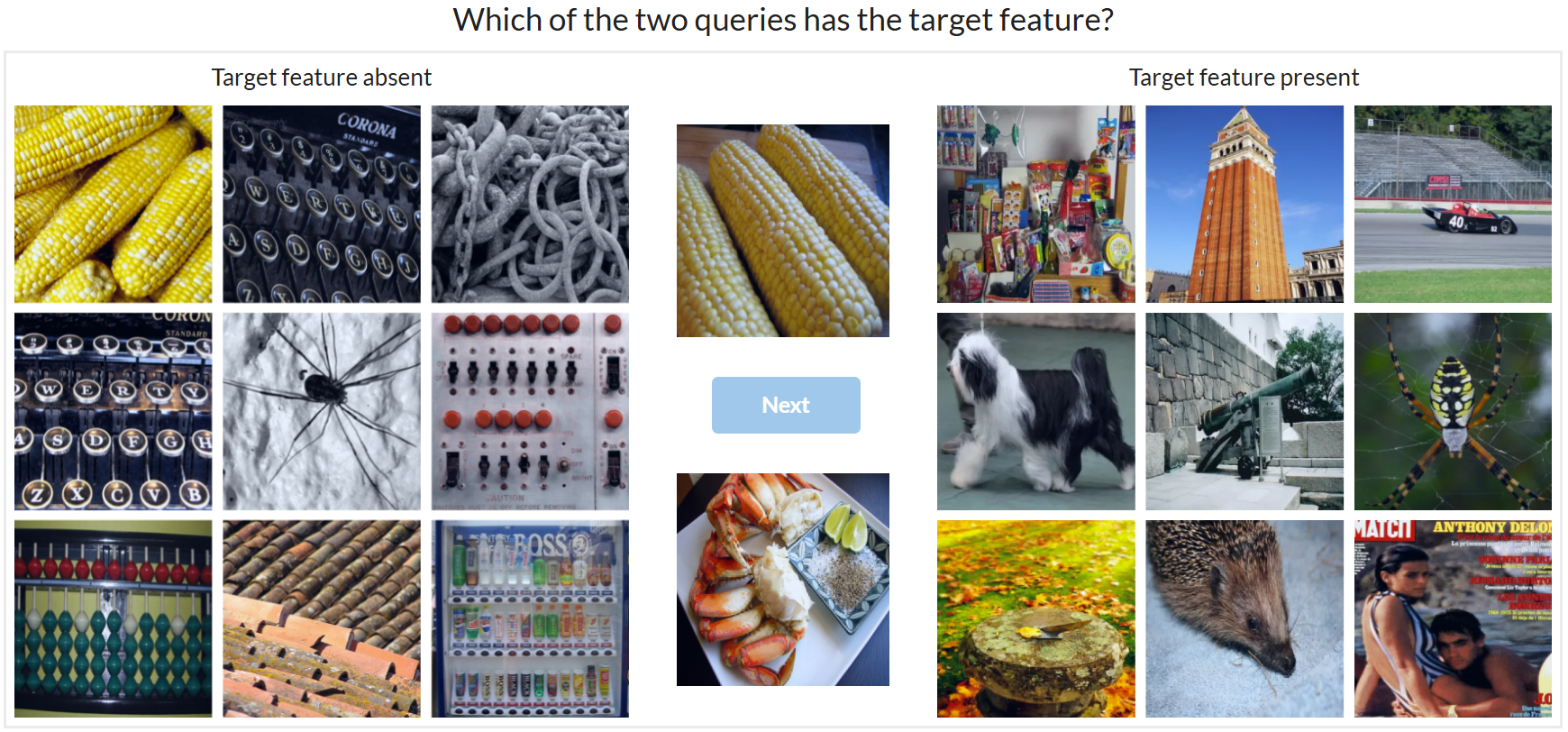}
  \caption{Instantiation of the trial from Fig.~\ref{fig:normal} in our control experiment (Sect.~\ref{sec:control}). In this protocol, we replace the images that highly activate the feature with images that carry little to no information about the feature's visual content. If the protocol only quantify the degree to which participant understand the feature, we expect them to perform at chance ($50\%$) in this setup.}\label{fig:control}
\end{figure}

\clearpage
\section{Scaling model interpretability}
\label{app:morefeatures}

Model-level interpretability is a property of a model's full feature population. Under a fixed annotation budget, we therefore face a choice: cover more features (\emph{breadth}) or measure each feature more carefully (\emph{depth}). This pilot asks which choice better approximates the population score.

\paragraph{What we do.} We collect interpretability judgments on $80$ DINOv2 neurons; this defines our reference score of $0.5555$. We then use bootstrapping ($200$ resamples per design) to simulate two cheaper alternatives: subsampling fewer units (breadth) or reducing the number of images and trials per unit (depth). For each design we report the bootstrap mean and SD.

\begin{table}[ht!]
\centering
\begin{tabular}{llcc}
\toprule
\textbf{Strategy} & \textbf{Setting} & \textbf{Bootstrap score} & \textbf{$\Delta$ vs. ref.} \\
\midrule
Reference & All $80$ units (full data) & $0.5555$ & --- \\
\midrule
Breadth & $10$ units & $0.5586 \pm 0.0291$ & $+0.0031$ \\
Breadth & $30$ units & $0.5541 \pm 0.0142$ & $-0.0014$ \\
Breadth & $60$ units & $0.5552 \pm 0.0057$ & $-0.0003$ \\
\midrule
Depth & $1$ image,  $1$ trial   & $0.5475 \pm 0.0376$ & $-0.0080$ \\
Depth & $1$ image,  $3$ trials & $0.5484 \pm 0.0278$ & $-0.0071$ \\
Depth & $10$ images, $1$ trial   & $0.5509 \pm 0.0083$ & $-0.0046$ \\
\bottomrule
\end{tabular}
\caption{\textbf{Breadth vs.\ depth under a fixed budget.} Bootstrap estimates of cheaper designs against the full-data reference. Breadth approaches the reference more quickly than depth.}
\label{tab:pilot_breadth_depth}
\end{table}

\paragraph{What we find.} Breadth converges quickly to the reference: $60$ units already match it within $\pm 0.006$. On the other hand, depth converges more slowly. Even $3$ trials on a single image ($\pm 0.0278$) is less stable than $30$ units alone ($\pm 0.0142$).

\paragraph{Why this points to scaling features.} Depth makes our estimate of the units we picked more precise; breadth makes it more representative. Because the $80$-unit reference is itself a stand-in for a much larger feature population, only breadth moves the estimate toward what we actually care about. We therefore allocate the main experiment's budget across as many features as possible rather than across repeated judgments on a small set of features.

\clearpage
\section{Overview of our feature extraction and explanation pipeline}\label{app:methodo}

\begin{figure*}[ht!]
\begin{center}
\centerline{
\includegraphics[width=\textwidth]{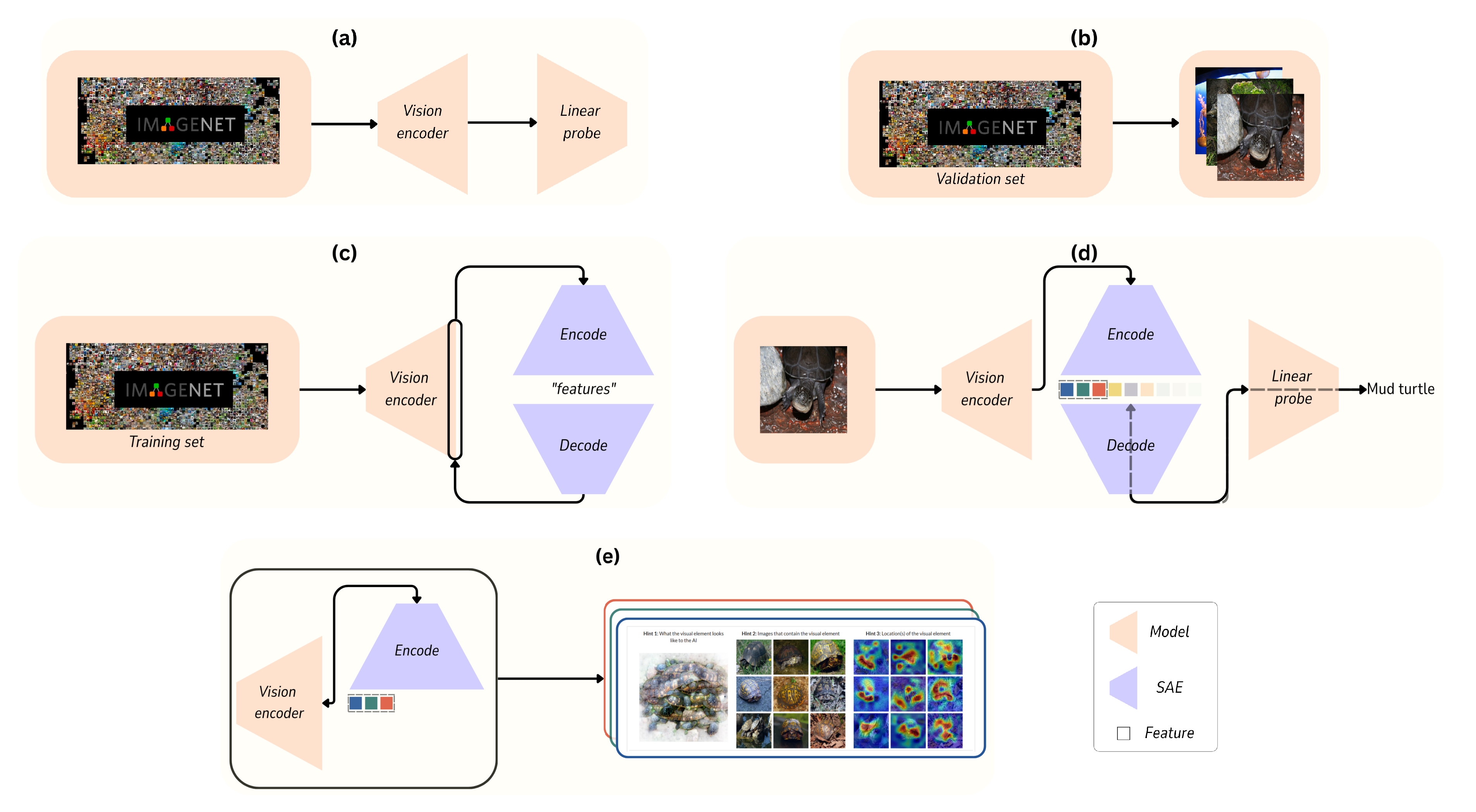}}
\caption{\textbf{Pipeline for extracting and explaining vision model features.}  (a) A linear probe is trained on ImageNet over the frozen backbone to derive per-feature importance and an accuracy baseline. (b) A representative subset of ImageNet images is selected to anchor evaluation across all models. (c) A TopK SAE is trained on patch-token activations to recover monosemantic features. (d) The most important features are selected per image, using Gradient×Input. (e) Each feature is explained via a MACO visualization~\cite{fel2023maco}, nine top-activating images, and RISE heatmaps~\cite{petsiuk2018rise}. See Section~\ref{sec:method_details} for full details.}
\label{fig:methodo}
\end{center}
\end{figure*}

\newpage

\section{SAE training, reconstruction, and stability}
\label{app:sae_quality}

\begin{table}[ht!]
\centering 
\caption{Reconstruction quality and dead-feature rates for the sparse autoencoder trained on each model.} \label{tab:sae_quality} \small 
\begin{tabular}{lcc} 
\toprule 
\textbf{Model} & \textbf{Reconstruction ($R^2$)} & \textbf{Dead features} \\ 
\midrule 
ViT-S & 0.7147 & 1.0\% \\ 
ViT-B & 0.7152 & 0.0\% \\ 
CLIP & 0.7257 & 2.5\% \\ 
DINOv3 & 0.8211 & 0.0\% \\ 
DINOv2 & 0.8019 & 0.0\% \\ 
SigLIP & 0.8963 & 0.0\% \\ 
\bottomrule 
\end{tabular} 
\end{table}

Reconstruction quality does not track human interpretability in the direction required for it to explain the supervised--foundation pattern. ViT-S and ViT-B have nearly identical reconstruction ($R^2=.7147$ and $.7152$) yet differ by 5.7 localizability points. Likewise, DINOv2 and DINOv3 differ only modestly in reconstruction but by approximately 9 localizability points. Dead-feature rates also do not explain the ordering: CLIP has the largest dead-feature fraction yet is the highest-scoring foundation model on semanticity.

\vspace{2mm}

We independently train a second SAE for each model using a different random seed. For every original feature, we compute its maximum cosine similarity to any feature in the second dictionary and treat similarities above $0.5$ as a match~\citep{thasarathan2025universal}.

\begin{table}[ht!] 
\centering 
\caption{Stability of SAE features across random seeds and its effect on median localizability. Stable features are those whose maximum cosine similarity with a feature in the independently trained dictionary exceeds $0.5$.} \label{tab:sae_stability} 
\small 
\begin{tabular}{lcccc} 
\toprule 
\textbf{Model} & \textbf{Stable features} & \textbf{All features} & \textbf{Stable only} & $\boldsymbol{\Delta}$ \\ 
\midrule 
ViT-S & 64.6\% & 0.803 & 0.802 & $-0.001$ \\ 
ViT-B & 80.4\% & 0.860 & 0.864 & $+0.004$ \\ 
CLIP & 50.1\% & 0.797 & 0.764 & $-0.033$ \\ 
SigLIP & 76.6\% & 0.714 & 0.743 & $+0.029$ \\ 
DINOv2 & 83.7\% & 0.710 & 0.698 & $-0.012$ \\ 
DINOv3 & 50.7\% & 0.796 & 0.791 & $-0.005$ \\ 
\bottomrule 
\end{tabular} 
\end{table}

Restricting the analysis to stable features changes model localizability by at most $.033$. The only ordering change is a swap between CLIP and DINOv3, whose original medians differ by only $.001$; both supervised models remain above all four foundation models.

\newpage
\section{Distribution of feature reuse across models}\label{sec:reuse}
\begin{table}[ht!]
\centering
\caption{Number of unique SAE features retained after filtering dense features for each model. Differences reflect varying degrees of feature reuse across architectures.}
\vspace{2mm}

\label{tab:reuse}
\small
\begin{tabular}{lcccccc}
\toprule
 & \multicolumn{2}{c}{\textbf{Supervised}} & \multicolumn{4}{c}{\textbf{Foundation}} \\
\cmidrule(lr){2-3}\cmidrule(lr){4-7}
 & ViT-S/16 & ViT-B/16 & DINOv2 & DINOv3 & CLIP & SigLIP \\
\midrule
\# Unique features & 1012 & 1305 & 1119 & 1018 & 545 & 896 \\
\bottomrule
\end{tabular}
\end{table}

\clearpage
\section{Psychophysics experiment}
\begin{figure}[ht!]
  \centering

  \begin{minipage}{0.48\textwidth}
    \centering
    \includegraphics[width=\linewidth]{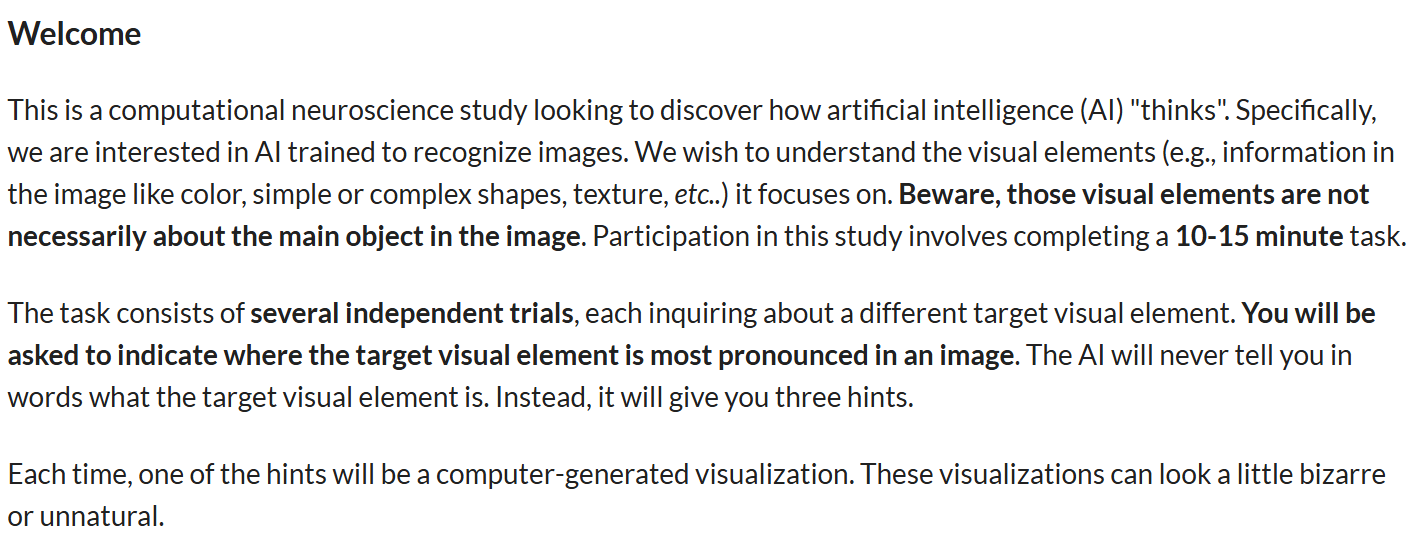}
    \caption*{\textbf{Localizability}}
  \end{minipage}
  \hfill
  \begin{minipage}{0.48\textwidth}
    \centering
    \includegraphics[width=\linewidth]{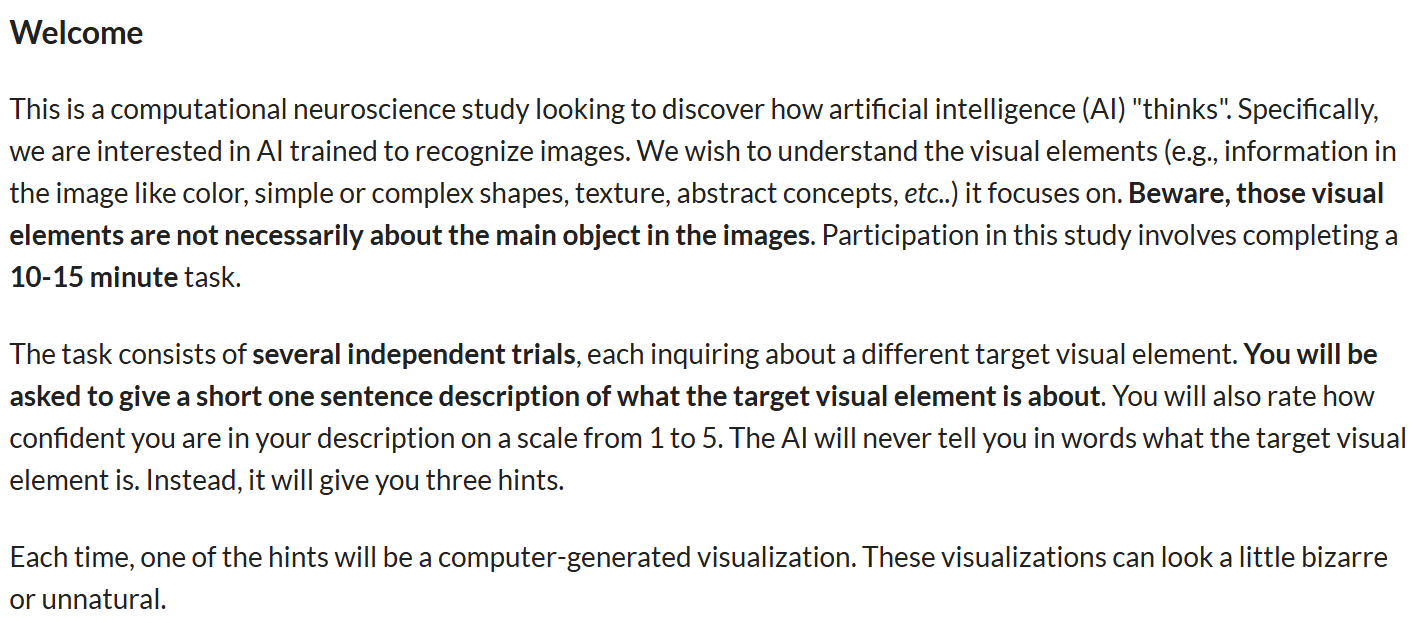}
    \caption*{\textbf{Semanticity}}
  \end{minipage}
  \caption{
  Welcome screens shown to participants before the psychophysics experiments.
  Left: instructions for the \emph{localizability} task.
  Right: instructions for the \emph{semanticity} task.
  }
  \label{fig:instructions}
\end{figure}

\begin{figure*}[ht!]
\begin{center}
\centerline{
\includegraphics[width=\textwidth]{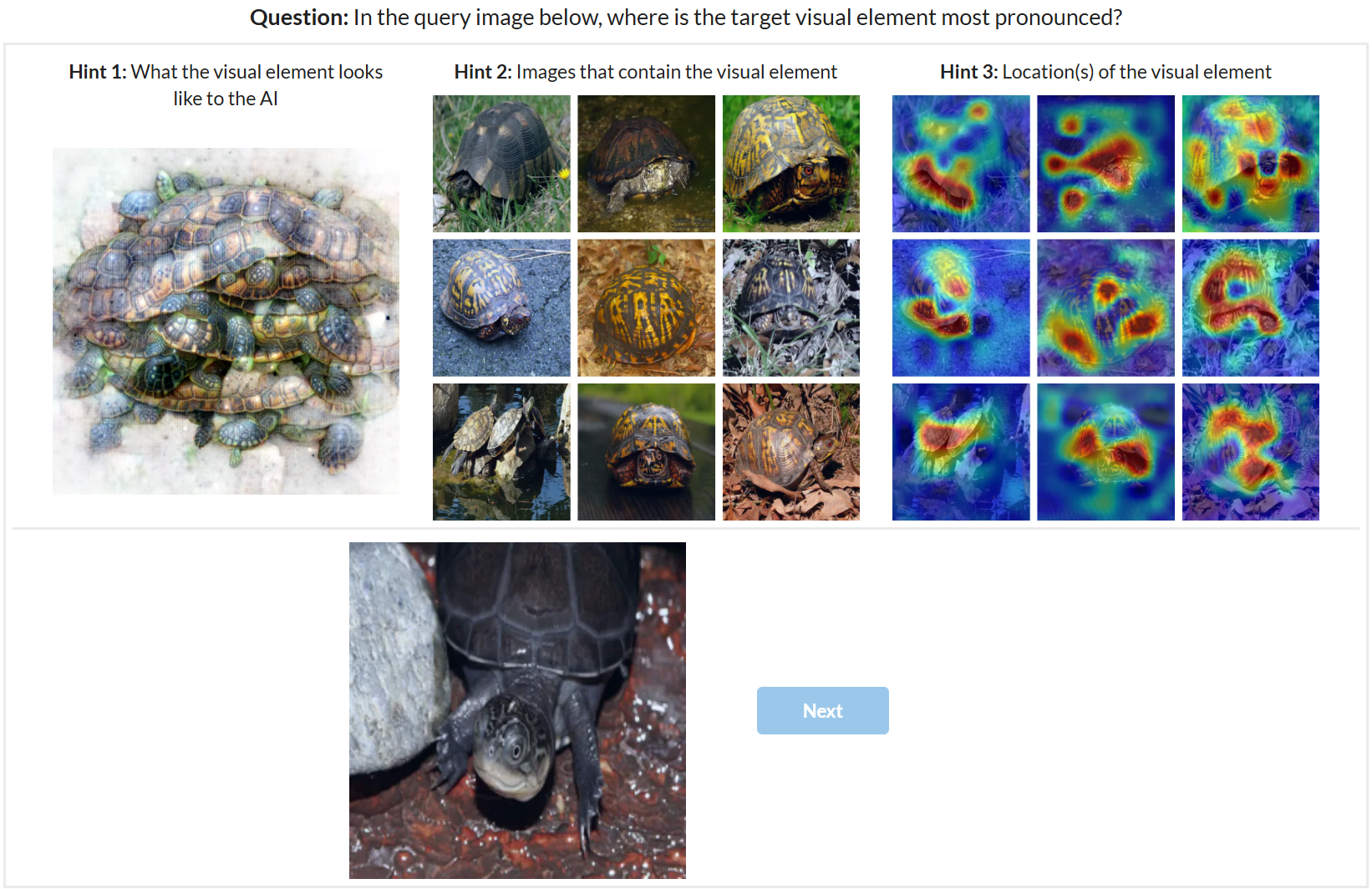}}
\caption{Screenshot of a trial for the \emph{localizability} experiment for DINOv3. A feature is explained through 3 panels at the top: (left) feature visualization by means of a maximally activating synthetic image, (middle) a set of images that highly activate the features, with their associated heatmaps (right) highlighting where the feature is located on those images. Participants were asked to click on a location where they expected the feature to be present on a new image (bottom). The more interpretable the feature, the more likely they are to correctly identify an area where the feature is present in the image.}
\label{fig:trial_loc}
\end{center}
\end{figure*}

\begin{figure*}[ht!]
\begin{center}
\centerline{
\includegraphics[width=\textwidth]{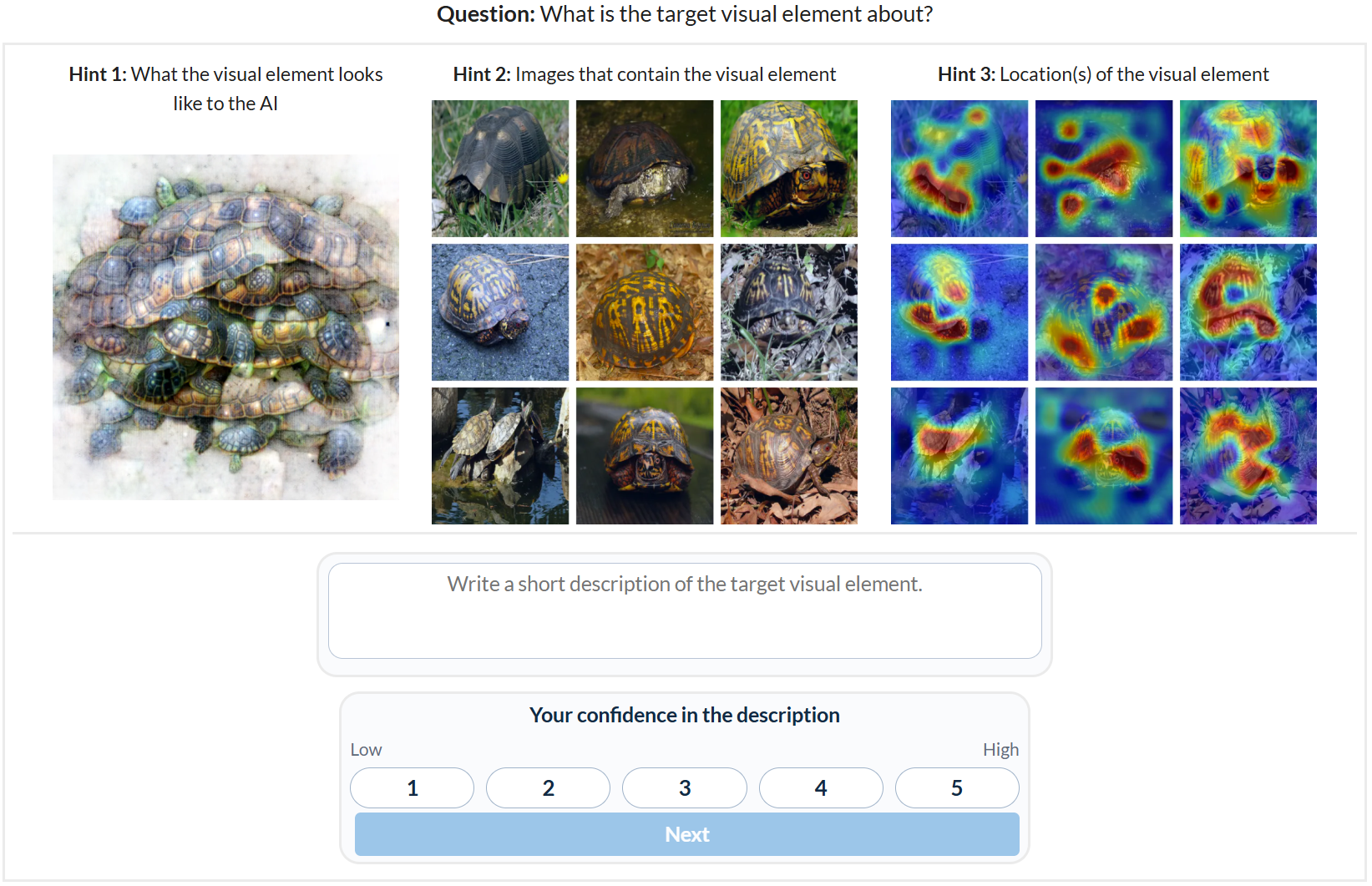}}
\caption{Screenshot of a trial for the \emph{semanticity} experiment for DINOv3. A feature is explained through 3 panels at the top: (left) feature visualization by means of a maximally activating synthetic image, (middle) a set of images that highly activate the features, with their associated heatmaps (right) highlighting where the feature is located on those images. Participants were asked to give a text-free description of what the feature is about (bottom). The more interpretable the feature, the more accurate their description.}
\label{fig:trial_name}
\end{center}
\end{figure*}

The experiment proceeded in three stages. Participants first read a welcome screen describing the task they were about to perform (Fig.~\ref{fig:instructions}). They then completed a short practice session to familiarize themselves with the task, followed by the main experiment. On each trial, a feature was presented through its visualization, top activating images, and corresponding heatmaps; depending on the condition, participants were asked either to click on the location where they expected the feature to appear in a new image (\emph{localizability}, Fig.~\ref{fig:trial_loc}) or to provide a free-text description of the feature (\emph{semanticity}, Fig.~\ref{fig:trial_name}).

\clearpage
\section{Quadrant of features}

\begin{figure*}[ht!]
\begin{center}
\centerline{
\includegraphics[width=0.9\textwidth]{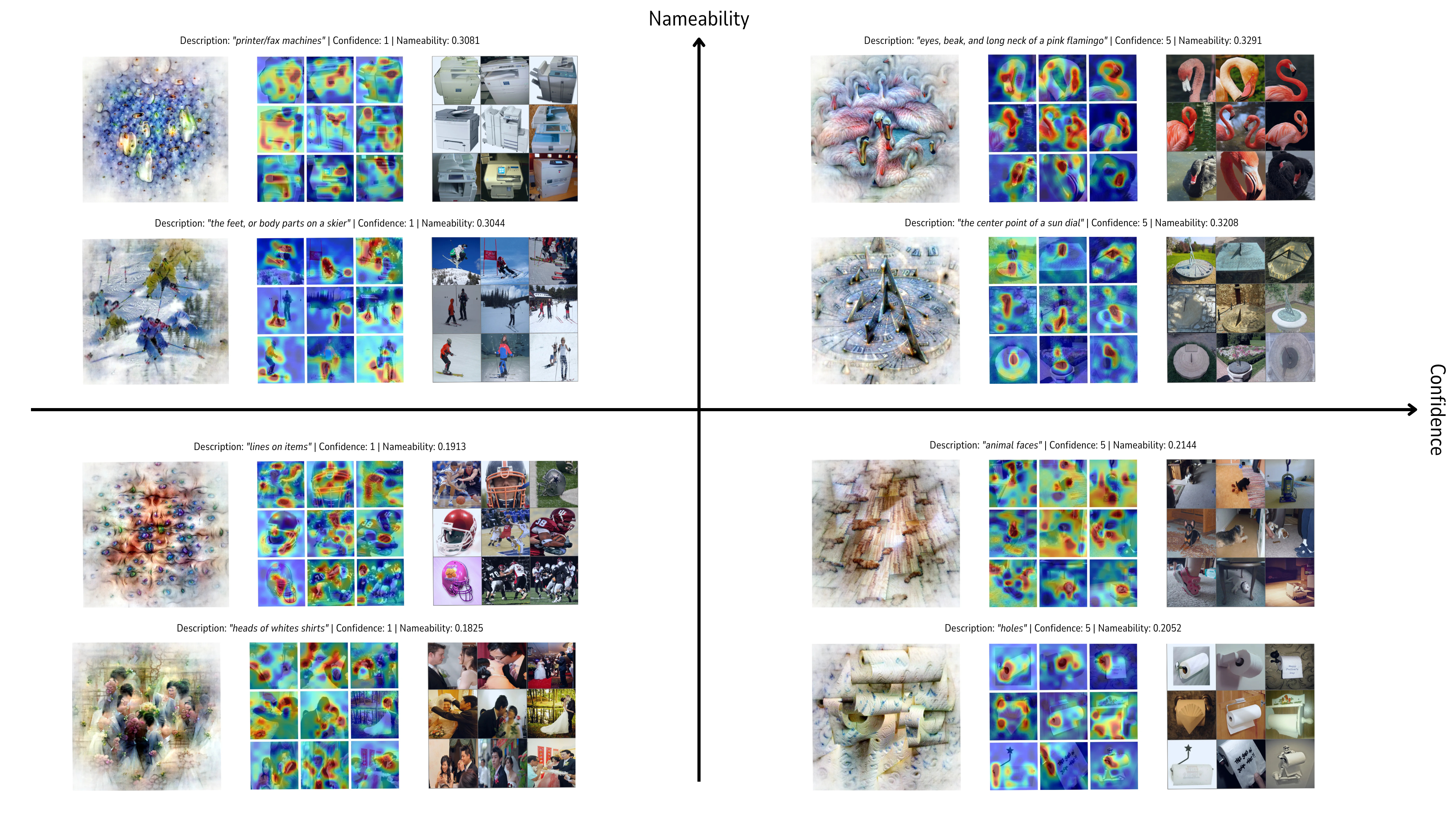}}
\caption{
\textbf{Relationship between \emph{semanticity} and confidence at the feature level.}
Examples of features from each of the four quadrants defined by human-rated confidence (x-axis) and \emph{semanticity} (y-axis). For each feature we show, from left to right: the feature visualization, heatmaps for top activating images, and the top activating images. The header above each example reports the human-provided description together with its confidence and \emph{semanticity} scores. Features in the upper-right quadrant are both confidently and accurately named, whereas features in the lower-left are neither confidently nor accurately named.}
\label{fig:quadrant_conf_name}
\end{center}
\end{figure*}

\begin{figure*}[ht!]
\begin{center}
\centerline{
\includegraphics[width=0.9\textwidth]{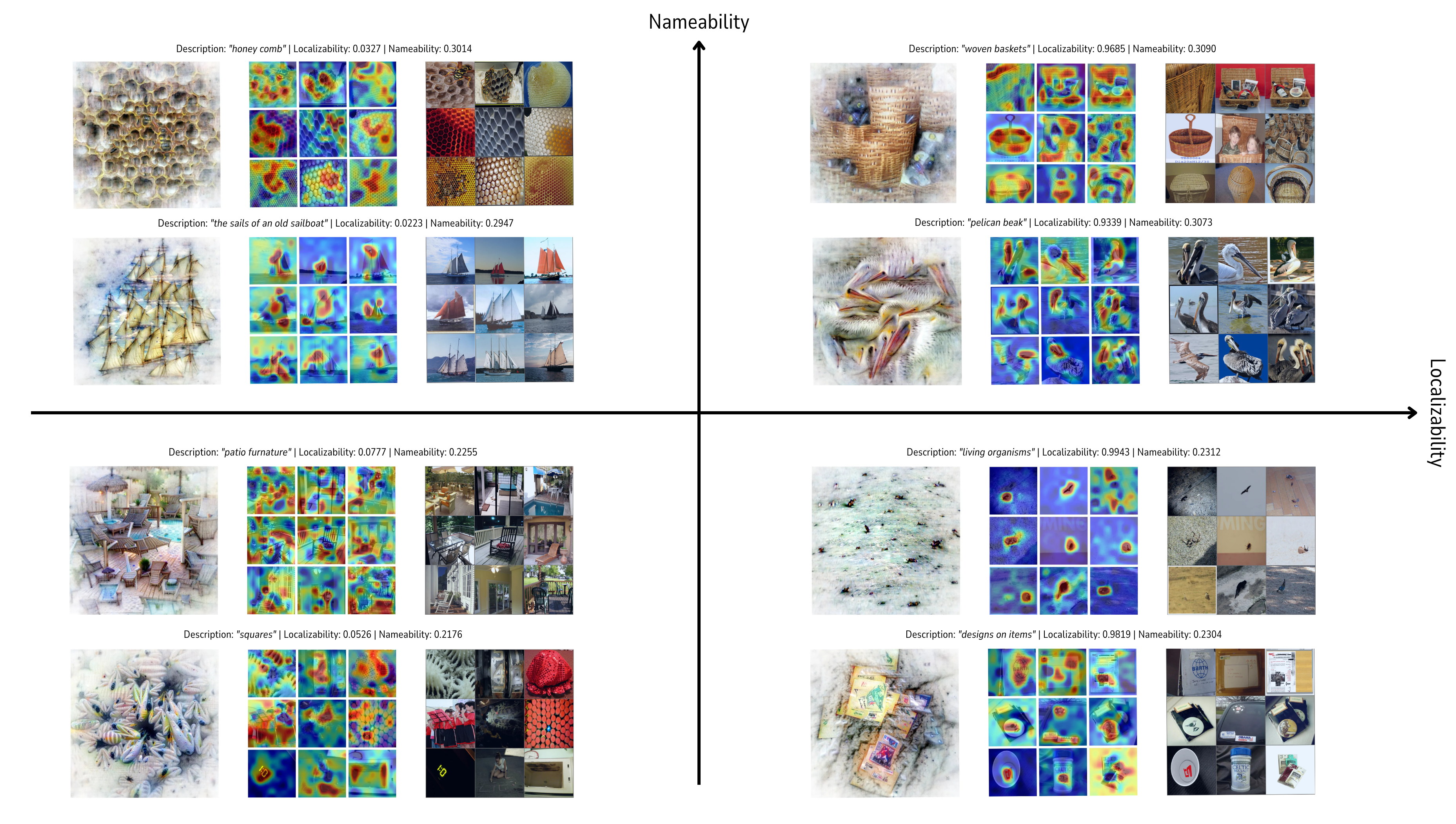}}
\caption{
\textbf{Relationship between \emph{semanticity} and \emph{localizability} at the feature level.}
Examples of features from each of the four quadrants defined by \emph{localizability} (x-axis) and \emph{semanticity} (y-axis). For each feature we show, from left to right: the feature visualization, heatmaps for top activating images, and the top activating images. The header above each example reports the human-provided description together with its \emph{localizability} and \emph{semanticity} scores. Features in the upper-right quadrant are both accurately localized and named, whereas features in the lower-left are not accurately localized or named.}
\label{fig:quadrant_loc_name}
\end{center}
\end{figure*}

\newpage
\subsection{Robustness to the choice of attribution method}
\label{app:hsic_robustness}

Our localizability score uses RISE heatmaps as a ground truth for evaluating participant clicks. To test whether the results depend on this choice, we repeat the analysis with HSIC~\citep{novello2022making}, an independent perturbation-based attribution method.

\paragraph{Agreement between RISE and HSIC.}
We first treat the maximum of each HSIC heatmap as a proxy participant click and score it against the corresponding RISE heatmap using the same \emph{localizability} as in the main experiment. HSIC peaks consistently fall within high-scoring regions of the RISE maps (median scores $0.889$--$0.949$). Scores differ across models (Kruskal--Wallis, $H=123.2$, $p<.001$), but the effect is negligible ($\epsilon^2=.012$), reflecting near-ceiling agreement throughout.

\paragraph{HSIC as ground truth.}
Then, we rescore the original participant clicks against HSIC rather than RISE heatmaps. Table~\ref{tab:hsic_rescoring} compares the resulting aggregate \emph{localizability}. Although absolute scores decrease under HSIC, the broad model ordering is retained and remains highly correlated with the original ranking (Spearman $\rho=.714$, $p=.111$, $n=6$).

\begin{table}[ht!]
\centering
\caption{\textbf{\emph{Localizability} under RISE and HSIC.}
\emph{Localizability} obtained by scoring the same participant clicks against heatmaps generated by either attribution method.}
\label{tab:hsic_rescoring}
\vspace{1mm}
\small
\begin{tabular}{lccc}
\toprule
\textbf{Model} & \textbf{RISE} & \textbf{HSIC} & $\boldsymbol{\Delta}$ \\
\midrule
ViT-S       & 0.803 & 0.692 & $-0.111$ \\
ViT-B       & 0.860 & 0.738 & $-0.122$ \\
DINOv2      & 0.710 & 0.649 & $-0.061$ \\
DINOv3      & 0.796 & 0.724 & $-0.072$ \\
CLIP        & 0.797 & 0.722 & $-0.075$ \\
SigLIP      & 0.714 & 0.578 & $-0.136$ \\
\bottomrule
\end{tabular}
\end{table}

Overall, these analyses indicate that the main localizability pattern is not specific to RISE.

\clearpage
\section{Plausibility robustness}\label{sec:semanticity_details}

\begin{figure}[ht!]
  \centering
\includegraphics[width=\textwidth]{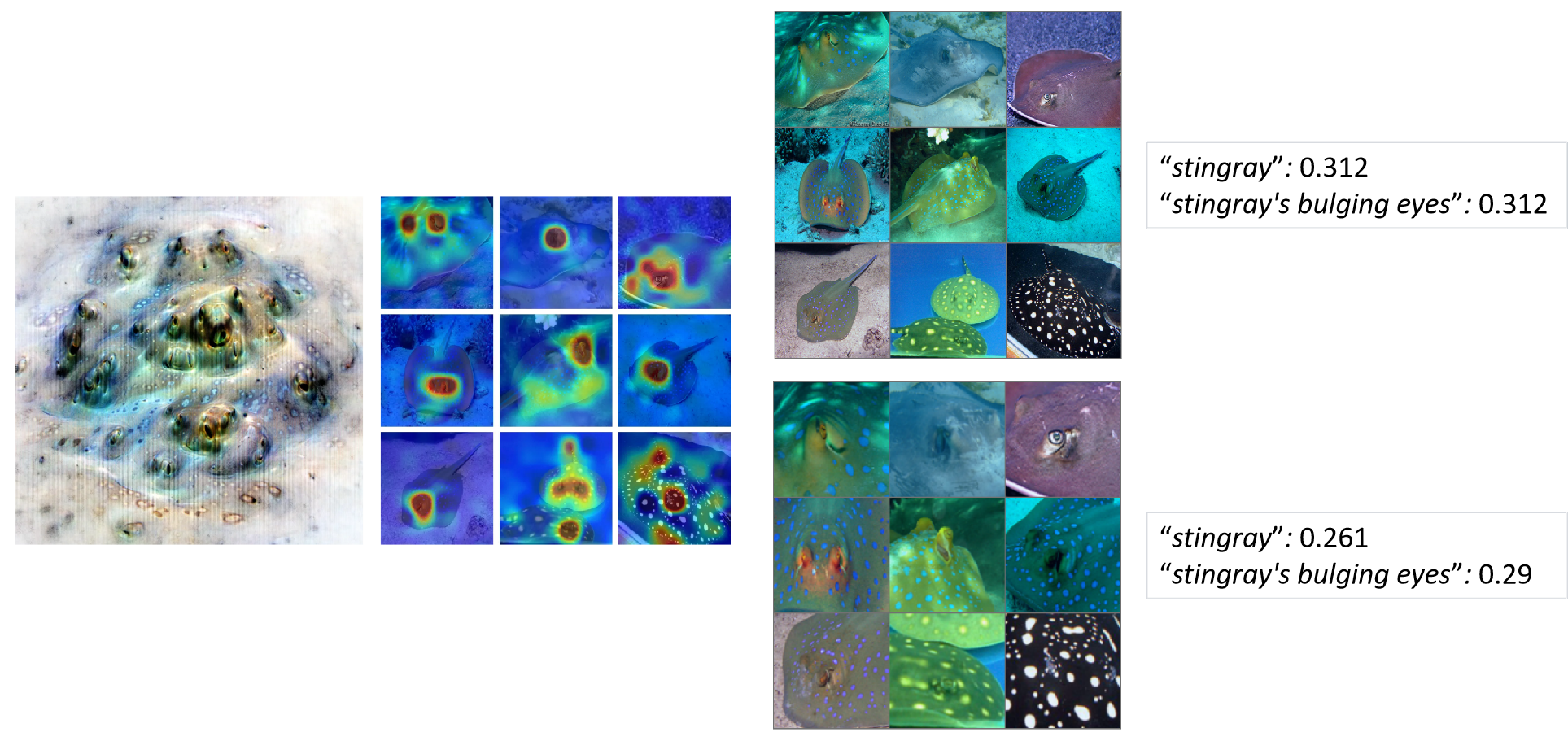}
  \caption{For a given feature, we evaluate the similarity of two descriptions, a general one consisting of the name of the main object, and a more specific one which we believe encapsulates the feature better. Using crops allows us to reward descriptions that go beyond simply naming the main object in the image.}\label{fig:crops}
\end{figure}

\paragraph{Crop vs full image.} For a given feature, we evaluate both the similarity of a general description---the name of the main object, and a more specific one which we believe encapsulates the feature better. Descriptions that go beyond simply naming the main object in the image generally receive a higher score when using crops (see Fig.~\ref{fig:crops} for an illustration).

Regardless, Table~\ref{tab:nameability_crop_full} highlights that results are fairly robust to the type of image used: \textit{full images vs. crops}. We also rescore the $4{,}470$ trials using $64$, $128$, and $160$ pixel crops in addition to the $96$ pixel crops reported in the main text. Relative to the $96$ pixel baseline, pooled trial-level correlations are $\rho=.910$, $.970$, and $.945$, respectively. The weakest per-model correlation is still $.843$ for CLIP ViT-B/16 at 64 pixels, and the model ranking remains unchanged.

\begin{table}[ht!]
\centering
\caption{Semanticity scores computed using the full image (bottom row) versus crops (top row) around the feature location. Best result per row is in bold, second best is underlined.}
\label{tab:nameability_crop_full}
\vspace{2mm}
\small
\begin{tabular}{lcccccc}
\toprule
 & \multicolumn{2}{c}{\textbf{Supervised}} & \multicolumn{4}{c}{\textbf{Foundation}} \\
\cmidrule(lr){2-3}\cmidrule(lr){4-7}
 & ViT-S/16 & ViT-B/16 & DINOv2 & DINOv3 & CLIP & SigLIP \\
\midrule
Crops ($\uparrow$)      & \textbf{0.274} & \underline{0.273} & 0.259 & 0.260 & 0.266 & 0.253 \\
Full image ($\uparrow$) & \underline{0.276} & \textbf{0.277} & 0.256 & 0.258 & 0.260 & 0.250 \\
\bottomrule
\end{tabular}
\end{table}

\paragraph{VLM scorer.} The main-text plausibility score is computed with CLIP ViT-B/16~\citep{radford2021learning}. To verify this choice does not drive the reported model ranking, we re-score every participant description with six additional vision-language encoders spanning different architectures and training objectives: CLIP ViT-L/14, OpenCLIP ViT-B/16 (LAION-2B), OpenCLIP ConvNeXt-B, SigLIP, SigLIP2, and BLIP (ITM). Each encoder is used to compute the same image-text cosine similarity score used for the main-text results ($n=4470$), and correlated (Spearman) against the CLIP ViT-B/16 score at the trial level. Table~\ref{tab:plausibility_vlm_robustness} reports the resulting per-backbone median scores and correlations. Four of the six alternative encoders closely replicate the main-text ranking ($\rho=.83$--$.94$, all $p<.001$); SigLIP2 and BLIP correlate slightly more weakly ($\rho=.74$--$.75$). Regardless, across all VLM scorers, among ViT-B, the supervised one stands at the top and SigLIP at the bottom.

\begin{table}[h]
\centering
\caption{\emph{Plausibility} scores per model, scored with CLIP ViT-B/16 (main text) and six alternative vision-language encoders, using the 9-crop feature-visualization images. $\rho$ is the trial-level Spearman correlation with the CLIP ViT-B/16 score ($n=4470$ trials, all $p<.001$). Bold is best, italic is second best, underline is worst.}
\label{tab:plausibility_vlm_robustness}
\begin{tabular}{lccccccc}
\toprule
Encoder & ViT-S & ViT-B & DINOv2 & DINOv3 & CLIP & SigLIP & $\rho$ \\
\midrule
CLIP ViT-B/16          & \textbf{0.274} & \textit{0.273} & 0.259 & 0.260 & 0.269 & \underline{0.253} & -- \\
\midrule
CLIP ViT-L/14                 & \textbf{0.225} & \textit{0.224} & 0.205 & 0.205 & 0.213 & \underline{0.194} & 0.907 \\
OpenCLIP ViT-B/16 (LAION-2B)  & \textit{0.238} & \textbf{0.239} & 0.222 & 0.221 & 0.226 & \underline{0.205} & 0.870 \\
OpenCLIP ConvNeXt-B           & \textbf{0.356} & \textit{0.354} & 0.337 & 0.338 & 0.344 & \underline{0.320} & 0.860 \\
SigLIP (ViT-B/16)             & \textbf{0.087} & \textit{0.082} & 0.045 & 0.043 & 0.045 & \underline{0.021} & 0.791 \\
SigLIP2 (SO400M)              & \textbf{0.061} & \textit{0.054} & 0.027 & 0.030 & 0.031 & \underline{0.012} & 0.706 \\
BLIP              & \textit{0.391} & \textbf{0.396} & 0.372 & 0.370 & 0.316 & \underline{0.264} & 0.699 \\
\bottomrule
\end{tabular}
\end{table}

\clearpage
\section{Detailed results for the psychophysics experiments.}\label{app:skew}
\begin{figure}[ht!]
    \centering

    \begin{subfigure}[t]{0.32\textwidth}
        \centering
        \includegraphics[width=\linewidth]{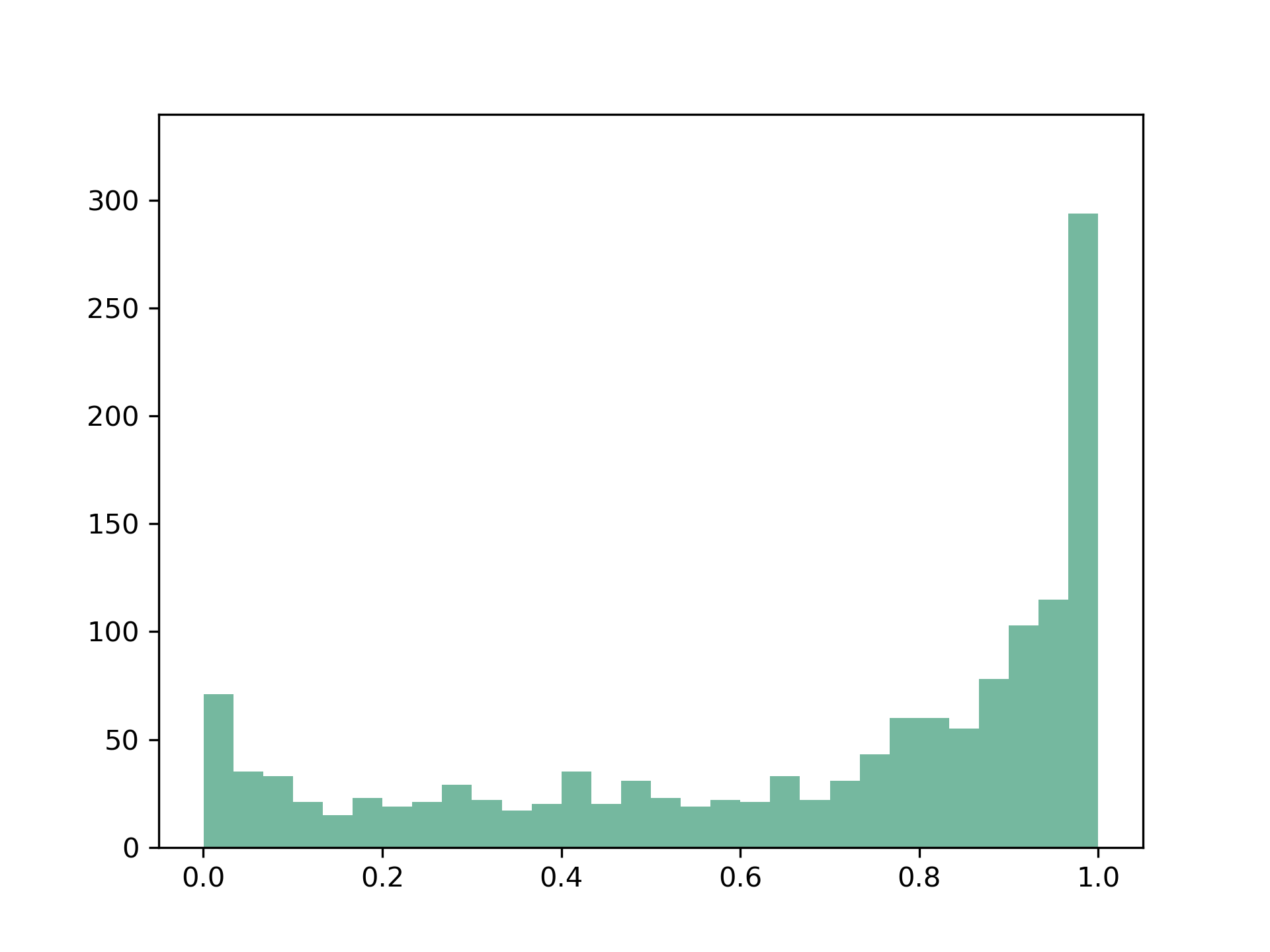}
        \caption{ViT-S}
    \end{subfigure}
    \hfill
    \begin{subfigure}[t]{0.32\textwidth}
        \centering
        \includegraphics[width=\linewidth]{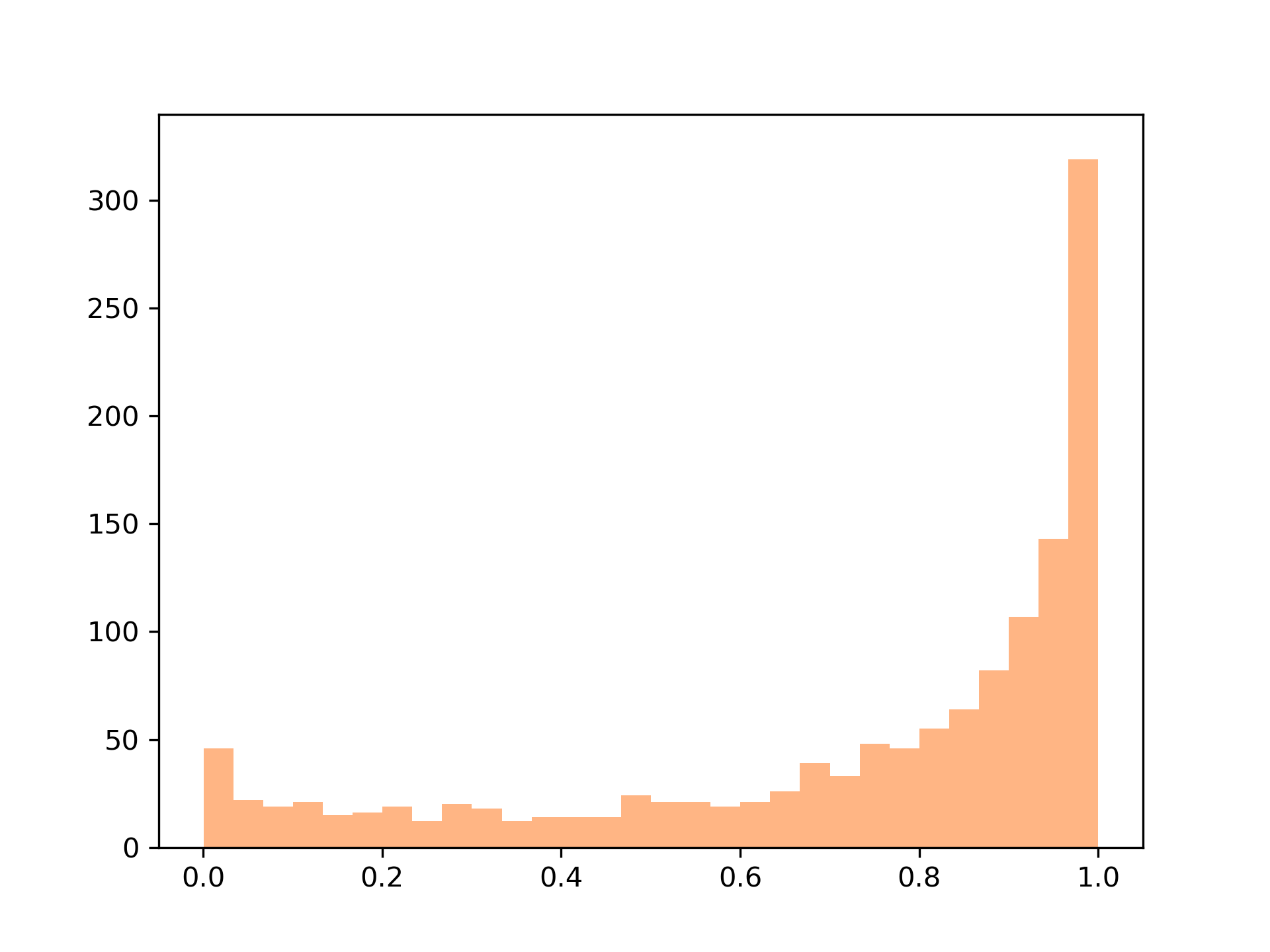}
        \caption{ViT-B}
    \end{subfigure}
    \hfill
    \begin{subfigure}[t]{0.32\textwidth}
        \centering
        \includegraphics[width=\linewidth]{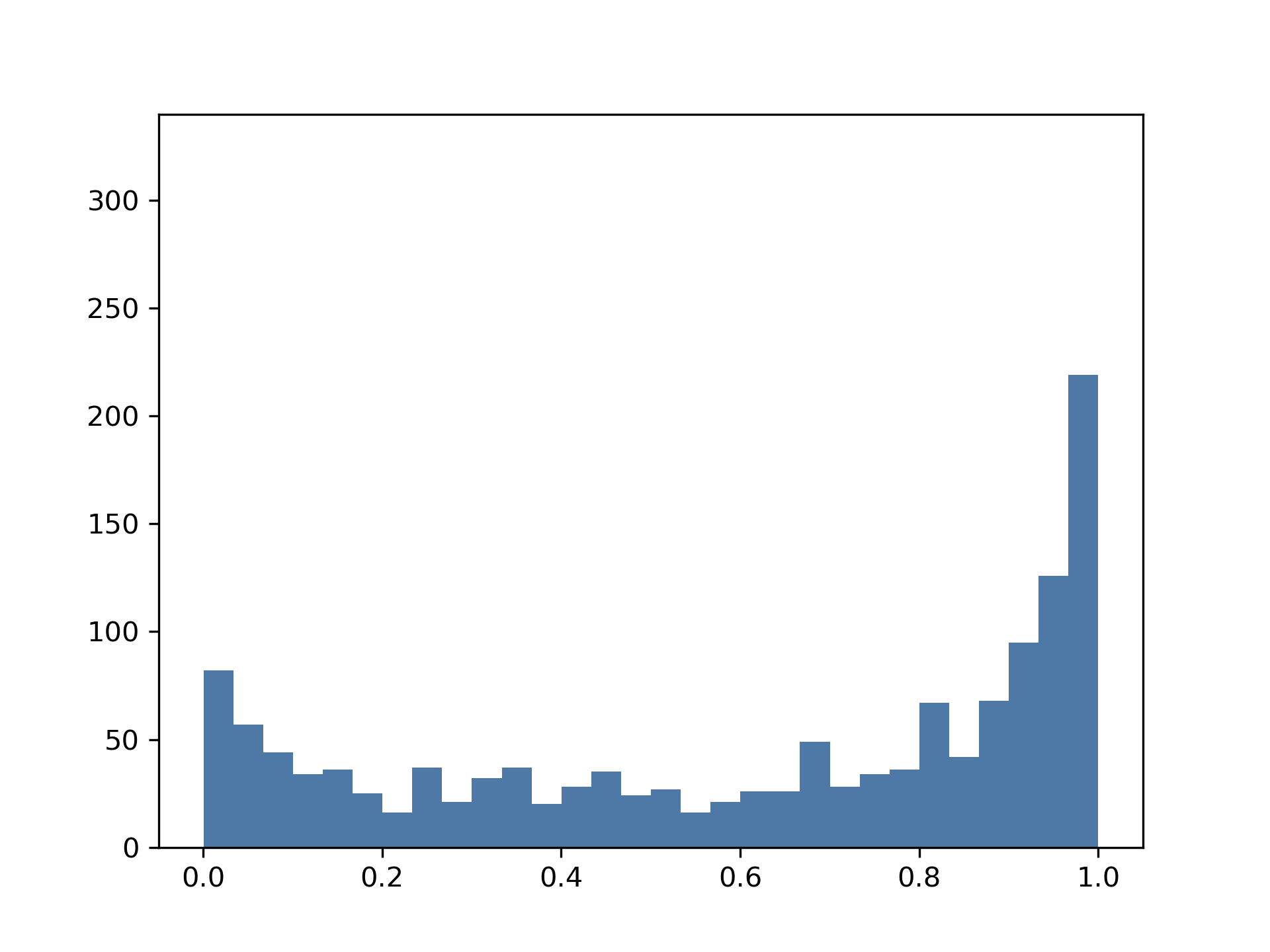}
        \caption{DINOv2}
    \end{subfigure}

    \vspace{2mm}

    \begin{subfigure}[t]{0.32\textwidth}
        \centering
        \includegraphics[width=\linewidth]{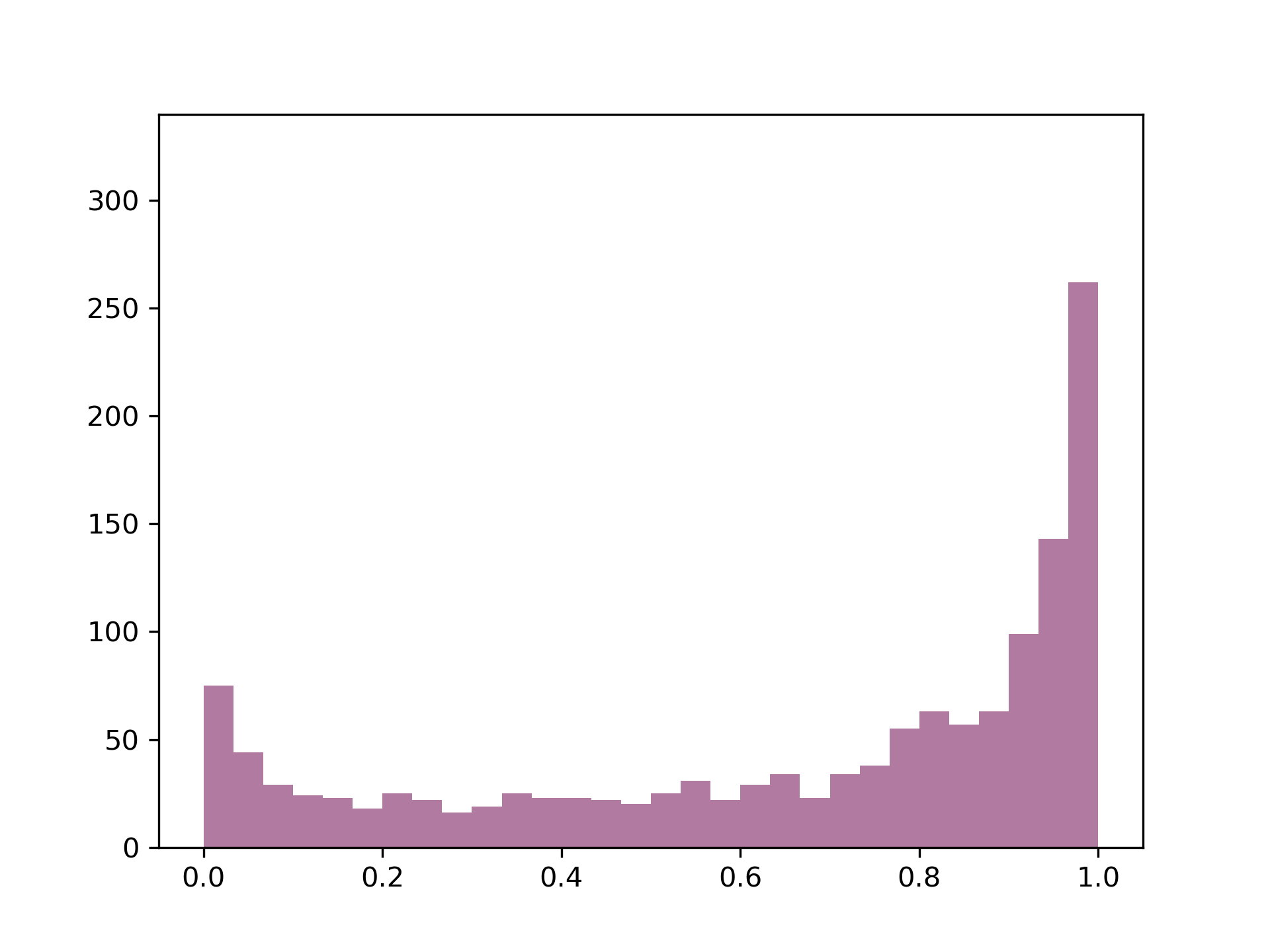}
        \caption{DINOv3}
    \end{subfigure}
    \hfill
    \begin{subfigure}[t]{0.32\textwidth}
        \centering
        \includegraphics[width=\linewidth]{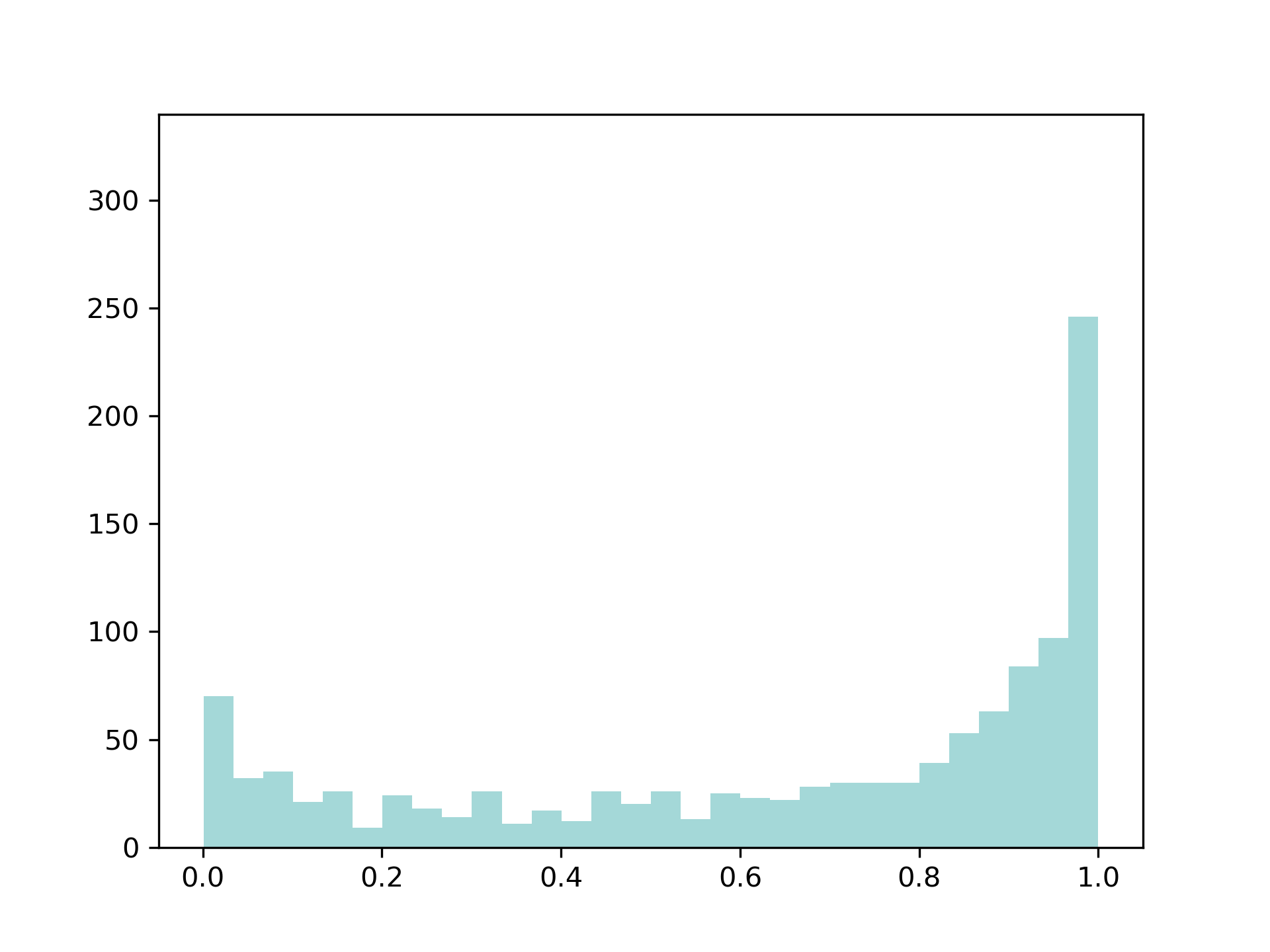}
        \caption{CLIP}
    \end{subfigure}
    \hfill
    \begin{subfigure}[t]{0.32\textwidth}
        \centering
        \includegraphics[width=\linewidth]{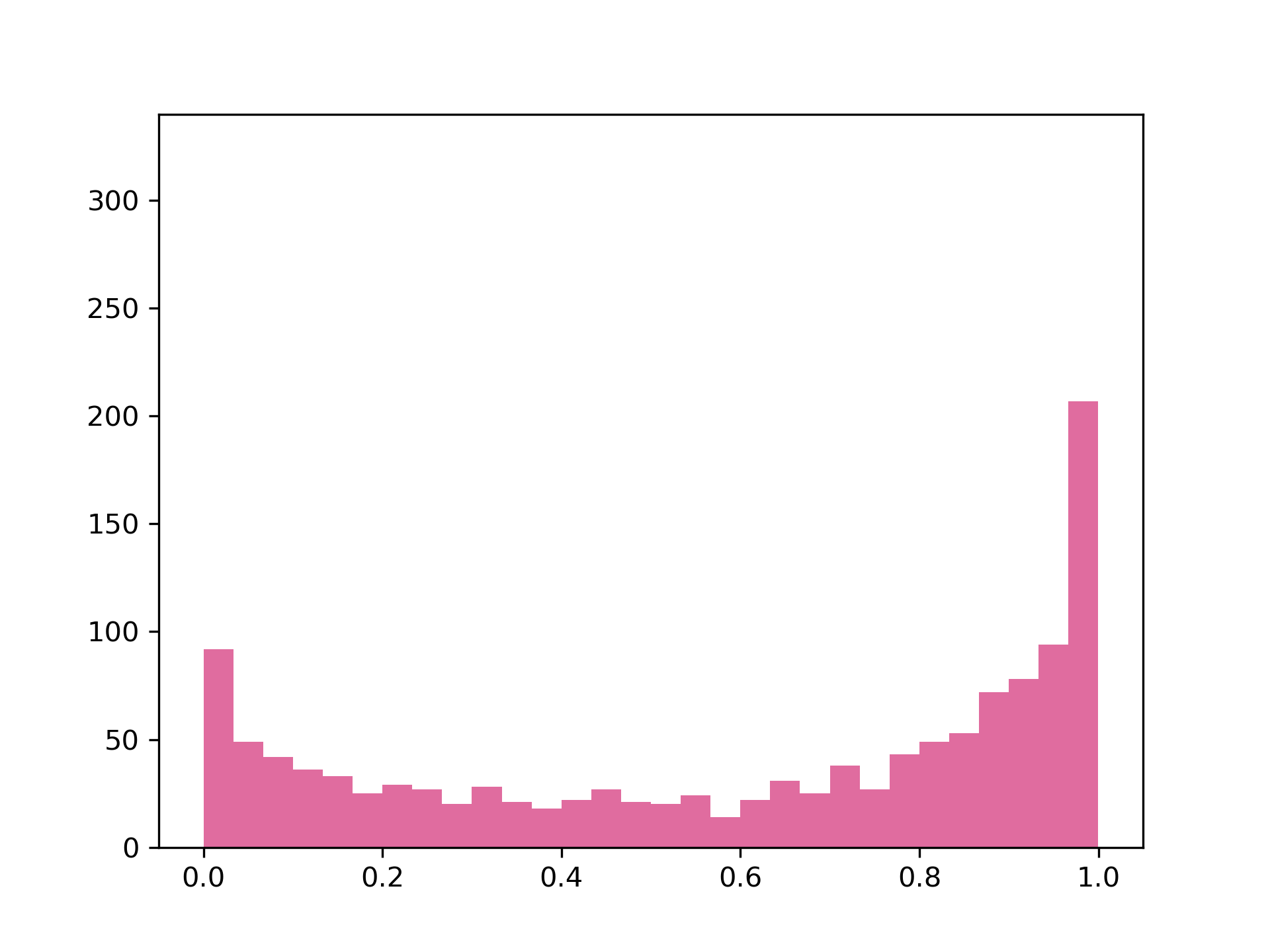}
        \caption{SigLIP}
    \end{subfigure}

    \caption{\textbf{Distribution of feature \emph{localizability} scores across models.} 
    Each panel shows the distribution of \emph{localizability} scores for one model.}
    \label{fig:distribution_localizability}
\end{figure}

\begin{figure}[ht!]
    \centering

    \begin{subfigure}[t]{0.32\textwidth}
        \centering
        \includegraphics[width=\linewidth]{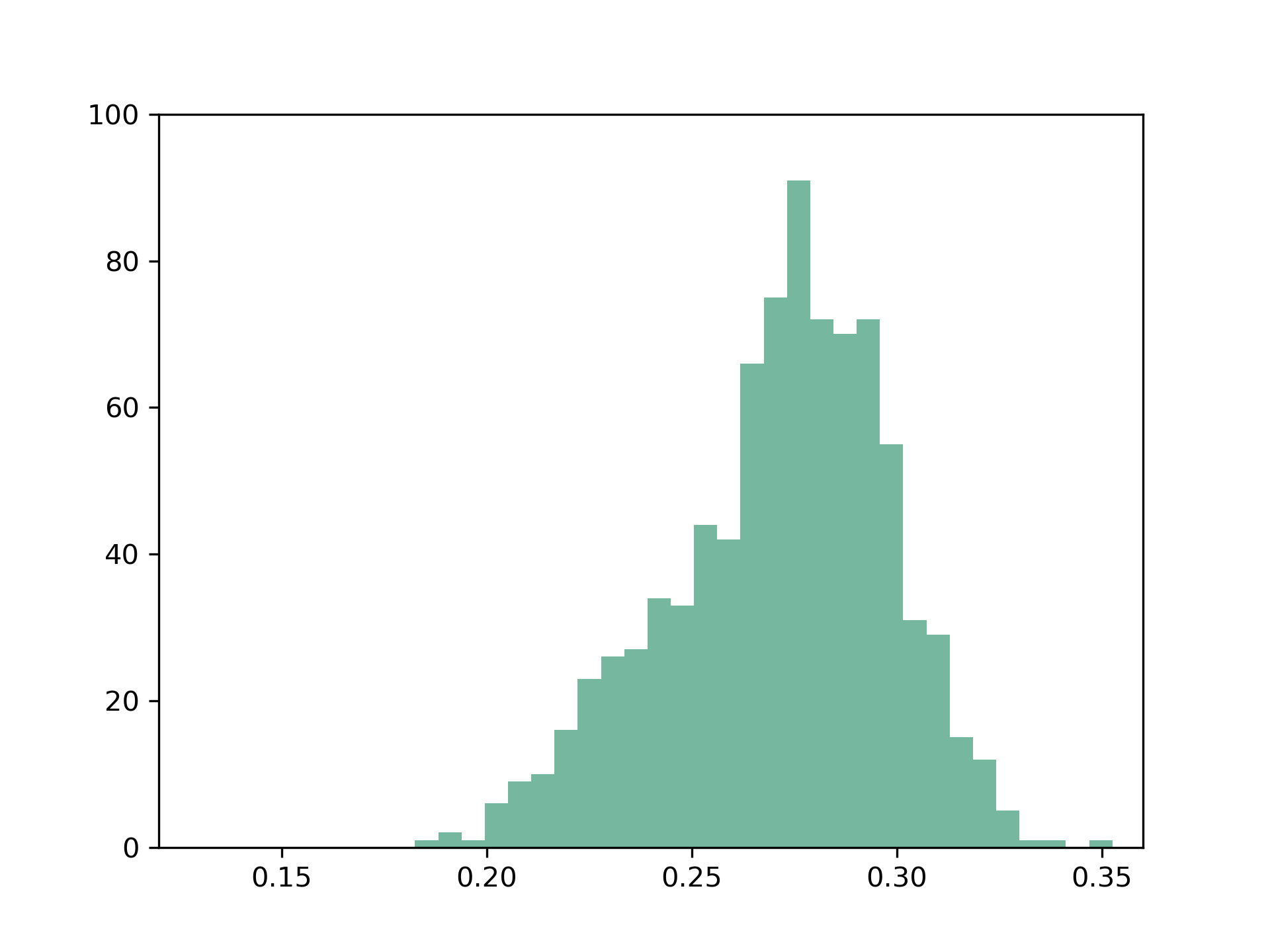}
        \caption{ViT-S}
    \end{subfigure}
    \hfill
    \begin{subfigure}[t]{0.32\textwidth}
        \centering
        \includegraphics[width=\linewidth]{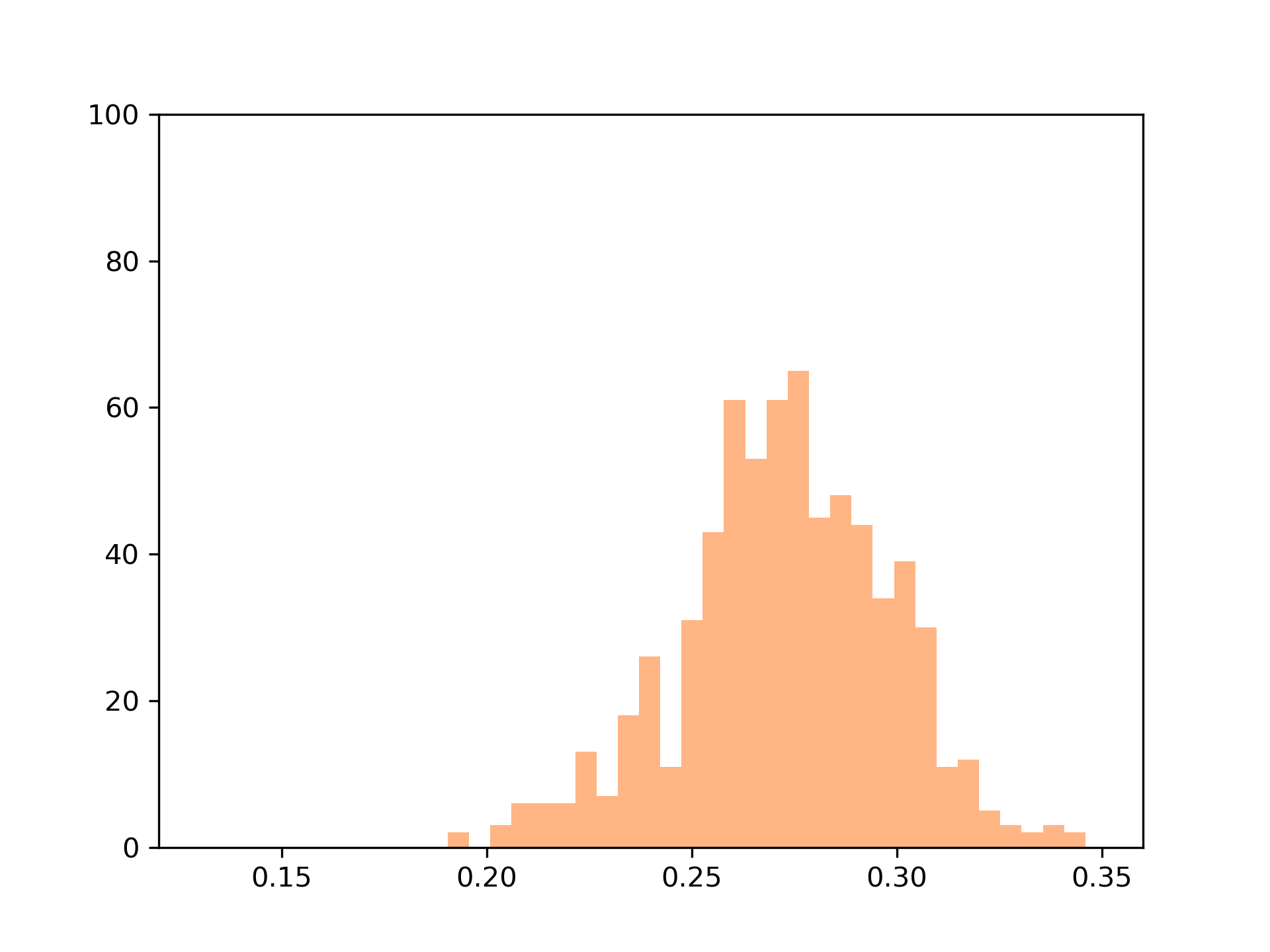}
        \caption{ViT-B}
    \end{subfigure}
    \hfill
    \begin{subfigure}[t]{0.32\textwidth}
        \centering
        \includegraphics[width=\linewidth]{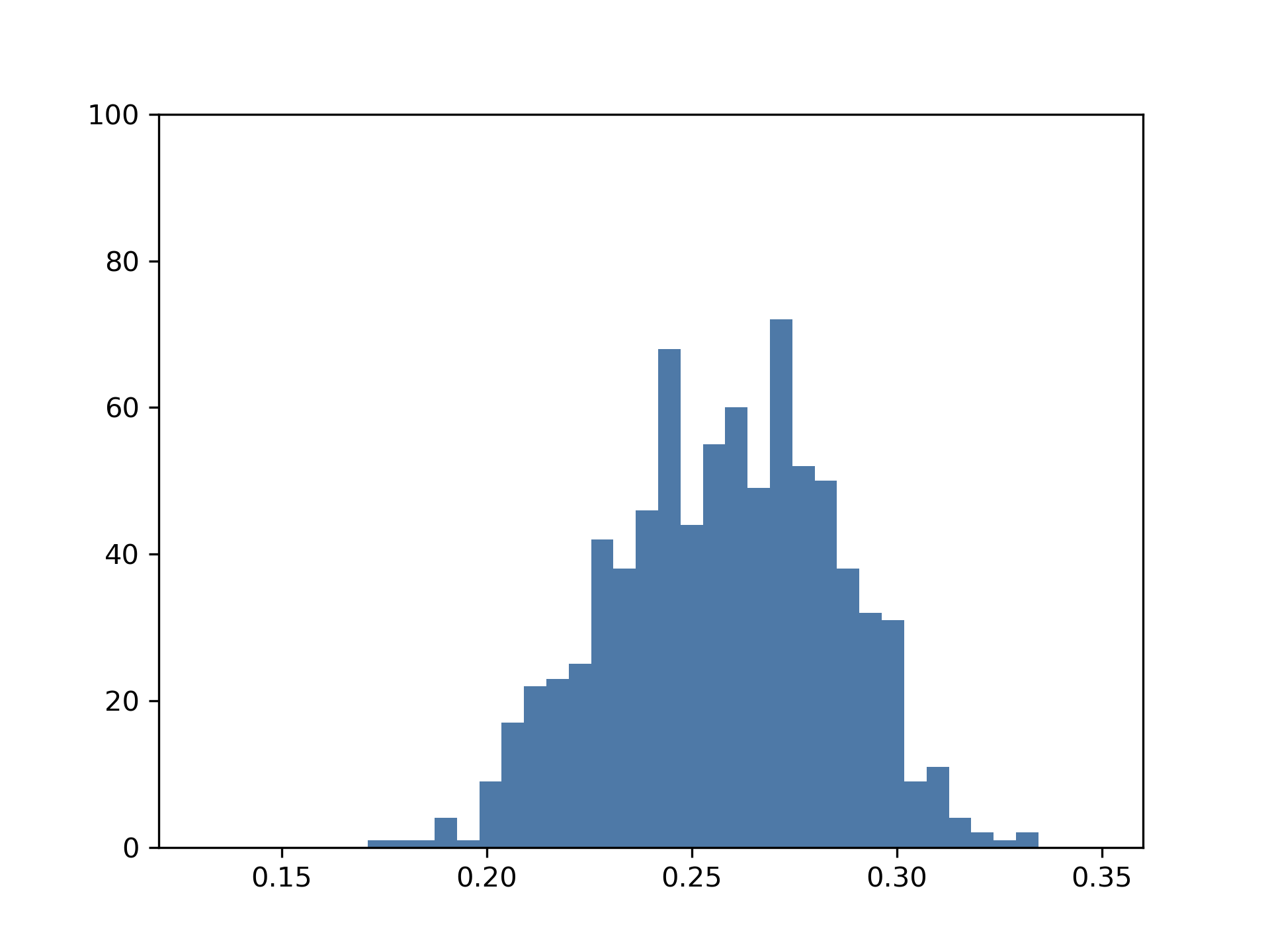}
        \caption{DINOv2}
    \end{subfigure}

    \vspace{2mm}

    \begin{subfigure}[t]{0.32\textwidth}
        \centering
        \includegraphics[width=\linewidth]{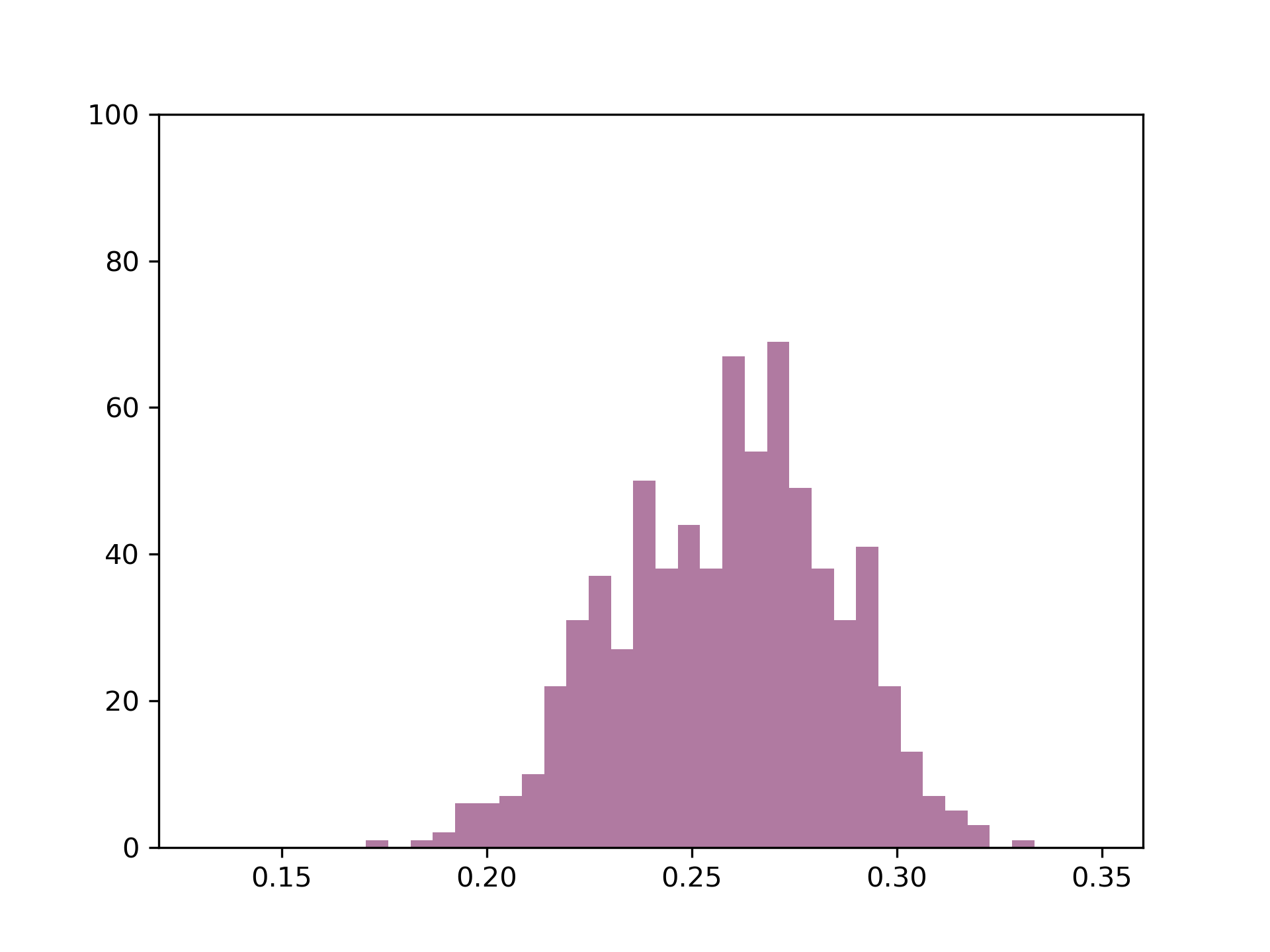}
        \caption{DINOv3}
    \end{subfigure}
    \hfill
    \begin{subfigure}[t]{0.32\textwidth}
        \centering
        \includegraphics[width=\linewidth]{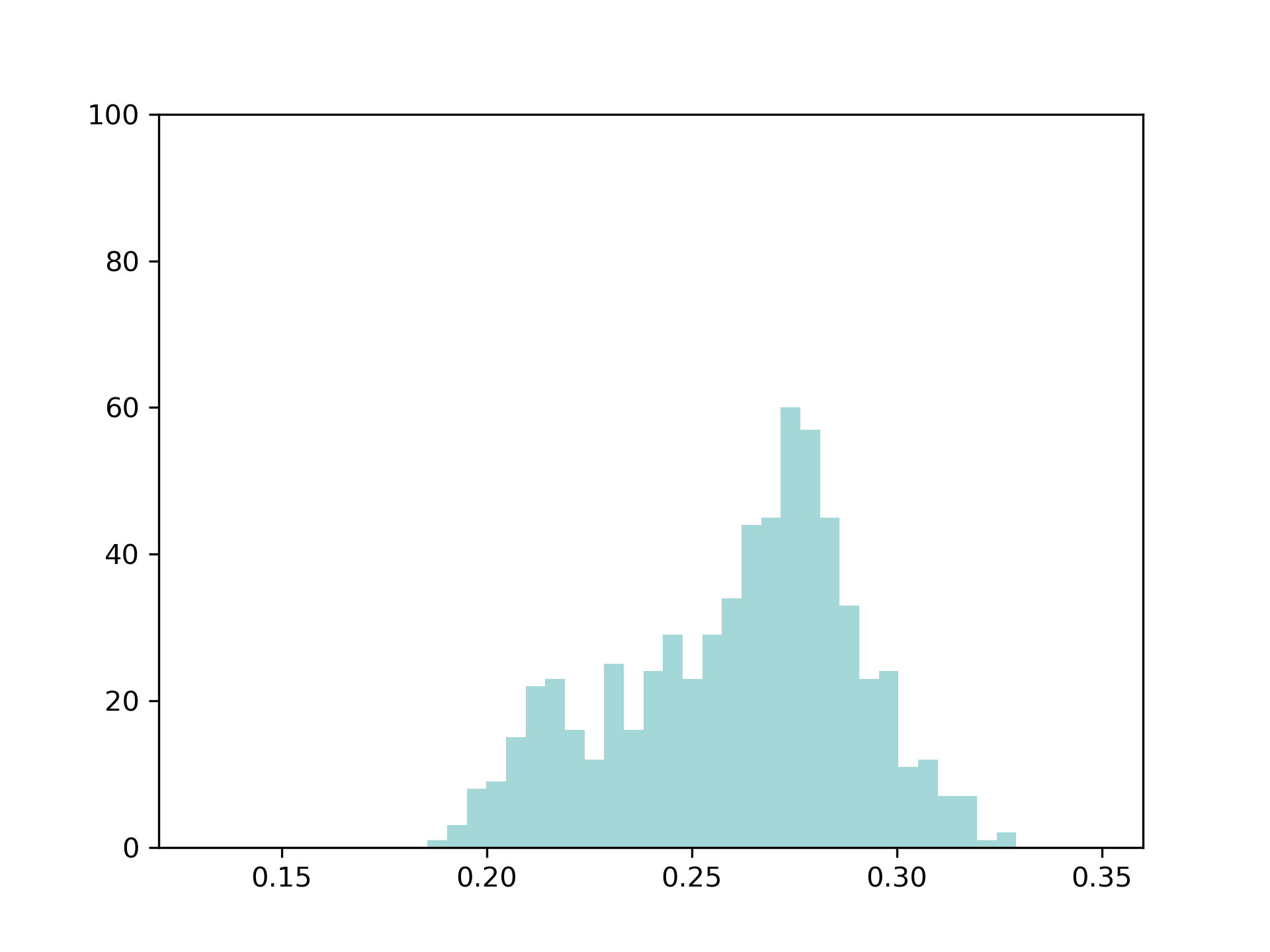}
        \caption{CLIP}
    \end{subfigure}
    \hfill
    \begin{subfigure}[t]{0.32\textwidth}
        \centering
        \includegraphics[width=\linewidth]{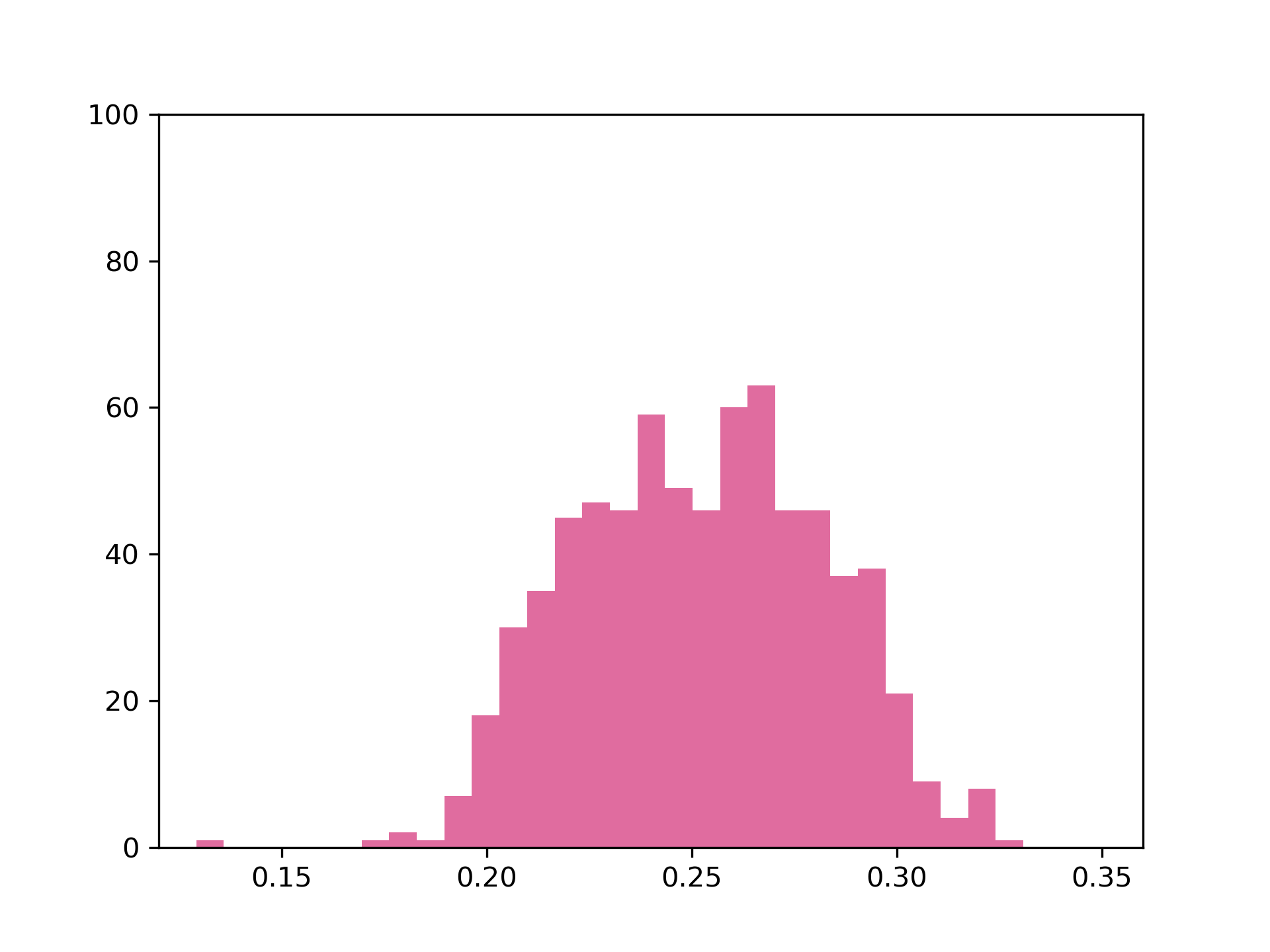}
        \caption{SigLIP}
    \end{subfigure}

    \caption{\textbf{Distribution of feature \emph{plausibility} scores across models.} 
    Each panel shows the distribution of \emph{plausibility} scores for one model.}
    \label{fig:distribution_nameability}
\end{figure}

Across both protocols, interpretability scores were non-normally distributed (Fig.~\ref{fig:distribution_localizability} and Fig.~\ref{fig:distribution_nameability}) for every model (Shapiro--Wilk test, $p < .001$). We therefore report the median score per model as a more representative measure of central tendency.

\begin{figure}[ht!]
    \centering

    \begin{subfigure}[t]{0.32\textwidth}
        \centering
        \includegraphics[width=\linewidth]{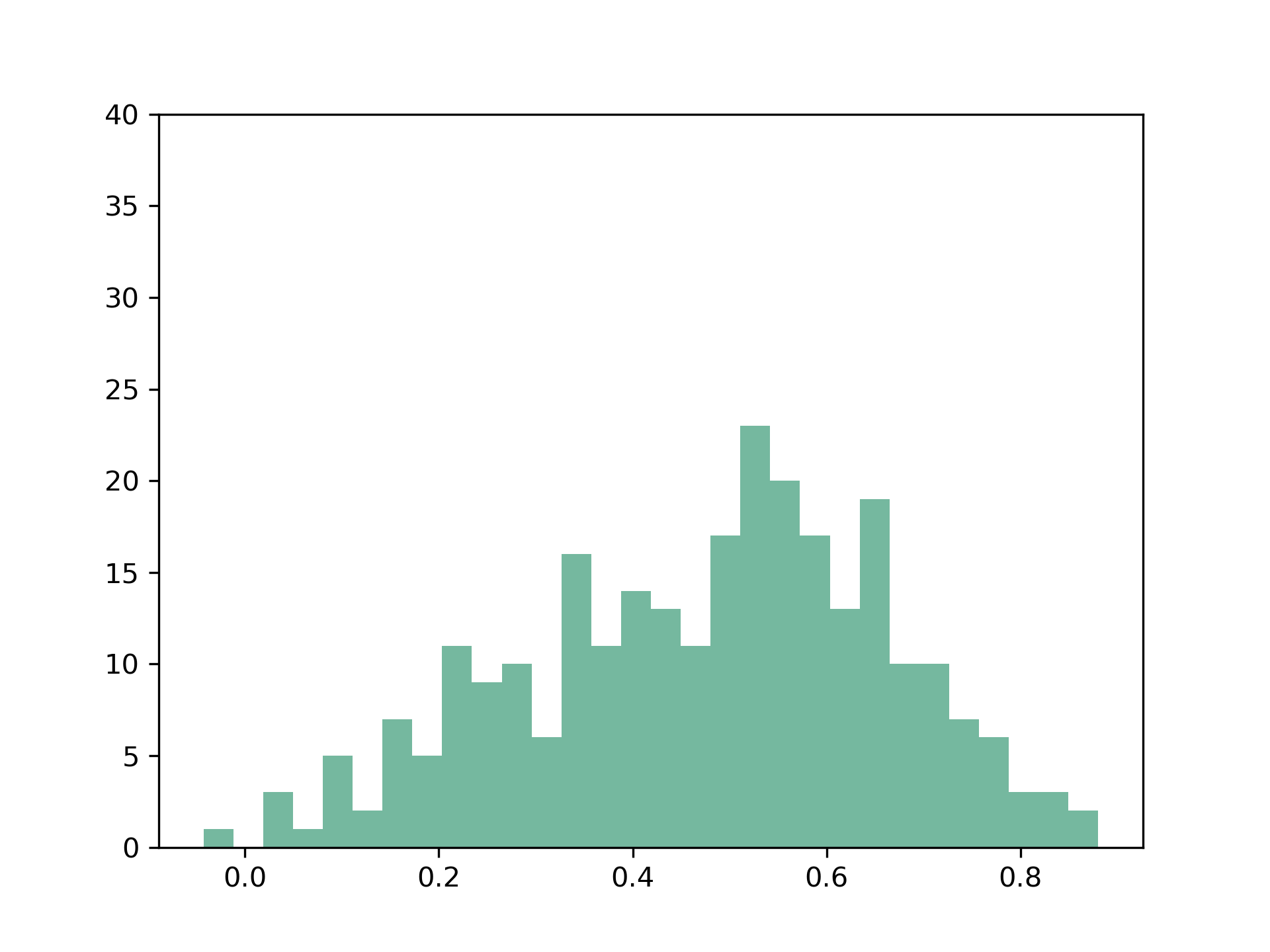}
        \caption{ViT-S}
    \end{subfigure}
    \hfill
    \begin{subfigure}[t]{0.32\textwidth}
        \centering
        \includegraphics[width=\linewidth]{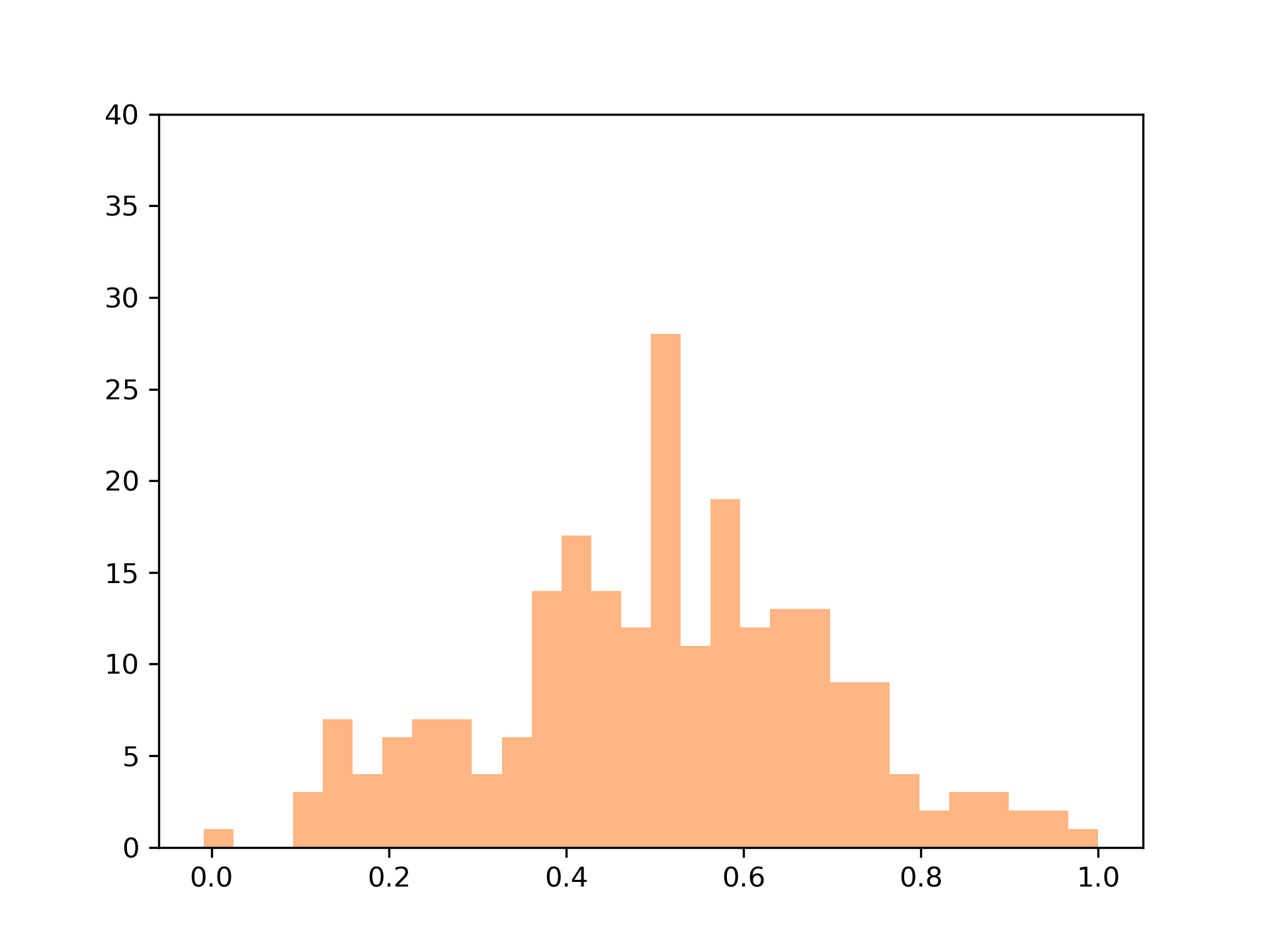}
        \caption{ViT-B}
    \end{subfigure}
    \hfill
    \begin{subfigure}[t]{0.32\textwidth}
        \centering
        \includegraphics[width=\linewidth]{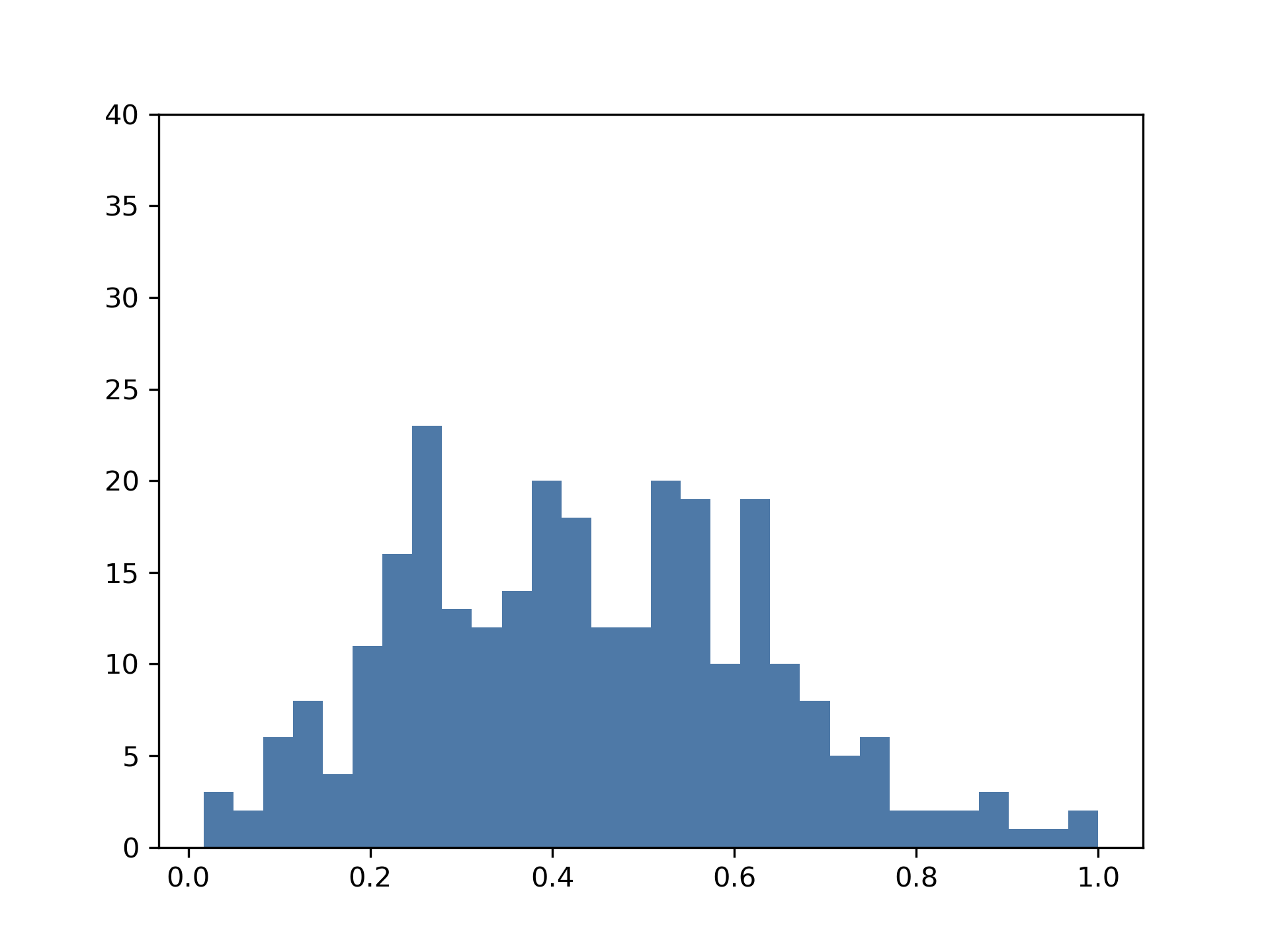}
        \caption{DINOv2}
    \end{subfigure}

    \vspace{2mm}

    \begin{subfigure}[t]{0.32\textwidth}
        \centering
        \includegraphics[width=\linewidth]{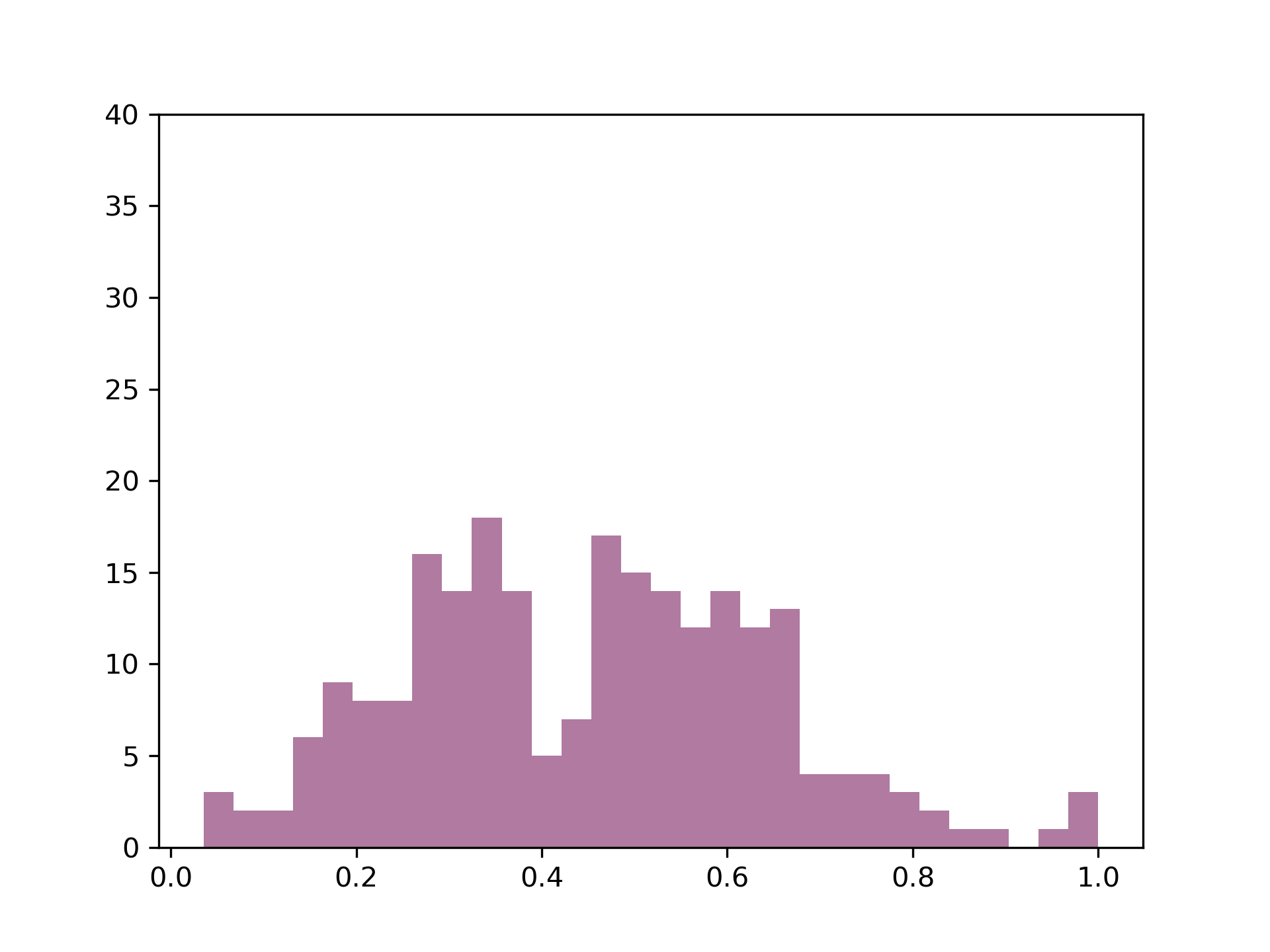}
        \caption{DINOv3}
    \end{subfigure}
    \hfill
    \begin{subfigure}[t]{0.32\textwidth}
        \centering
        \includegraphics[width=\linewidth]{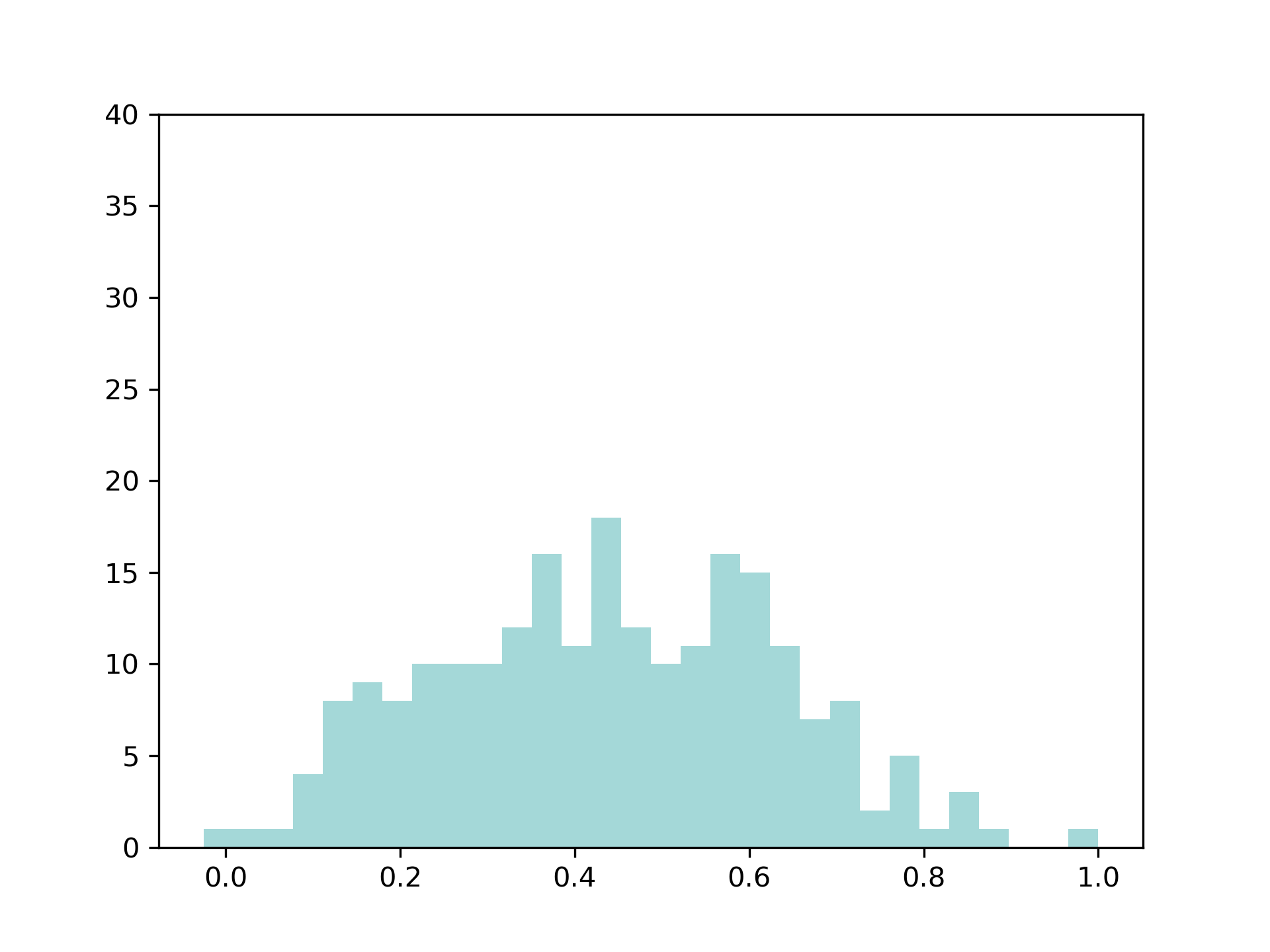}
        \caption{CLIP}
    \end{subfigure}
    \hfill
    \begin{subfigure}[t]{0.32\textwidth}
        \centering
        \includegraphics[width=\linewidth]{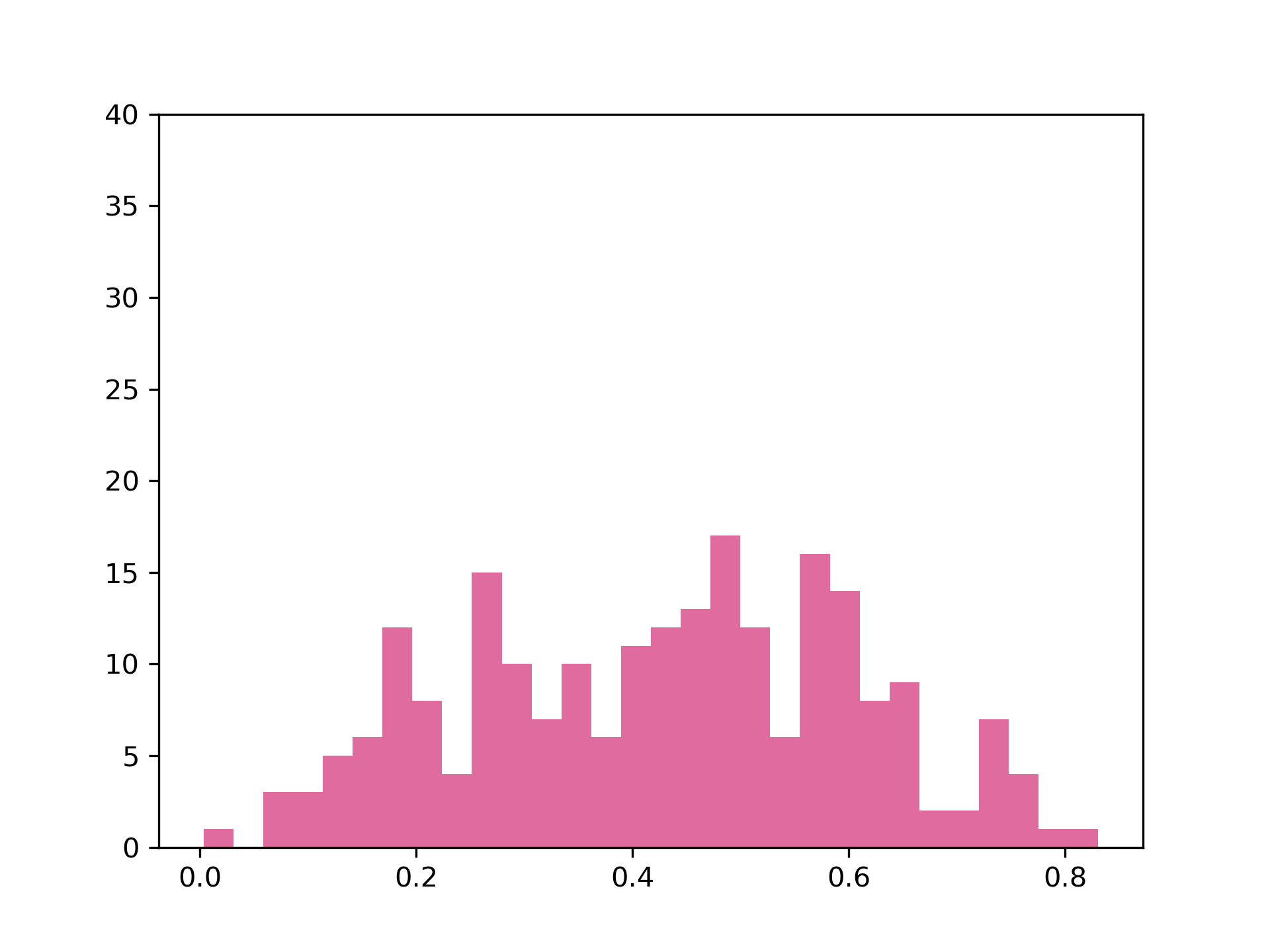}
        \caption{SigLIP}
    \end{subfigure}

    \caption{\textbf{Distribution of feature \emph{agreement} scores across models.} 
    Each panel shows the distribution of \emph{agreement} scores for one model.}
    \label{fig:distribution_agreement}
\end{figure}

\clearpage
\section{Additional results on locality of representations}
\label{app:kolmogorov}

A feature can be considered local if it fires on a few grouped pixels, i.e., (i) \emph{few} pixels are active and (ii) those active pixels are \emph{spatially grouped}. The Hoyer sparsity used in Section~\ref{sec:sparsity} primarily targets the first component. As a complementary proxy for the latter, we use the \textbf{Kolmogorov complexity}~\cite{li2004similarity} of each feature's RISE heatmaps, approximated by their JPEG compressibility: the size of the losslessly compressed heatmap, normalized by its uncompressed size. Grouped activations form contiguous, redundant regions and compress well; scattered activations do not. Lower values therefore indicate more spatially grouped heatmaps. We average over the heatmaps of each feature's nine most activating images and then over all features for a model.

\paragraph{Results.} Across the six models, compressibility correlates across all three interpretability measures in the expected direction---interpretable models tend to have more compressible heatmaps---although only significantly for \emph{agreement}: \emph{localizability} ($\rho = -0.64$, $p = 0.169$), \emph{plausibility} ($\rho = -0.52$, $p = 0.289$), and \emph{agreement} ($\rho = -0.88$, $p = 0.021$).
% (see Fig.~\ref{fig:compressibility})

\paragraph{Limitation.} Compressibility alone does not distinguish a heatmap with a single active pixel and a heatmap where every pixel takes the same value, as both are highly compressible. The score, therefore, captures the presence and, to some extent, the number of activation groups, but not whether those groups are spatially confined. We therefore read it as a useful complement to Hoyer sparsity: suggestive of grouping, but, on its own, an incomplete proxy for locality.

% \begin{figure}[ht!]
%     \centering
%     \includegraphics[width=\linewidth]{figures/compressibility.png}
%     \caption{\textbf{Compressibility of feature heatmaps as a complementary measure of locality.} Each panel plots an interpretability score against the mean compressibility of a model's feature heatmaps (lower $=$ more grouped activations). Left: \emph{localizability}. Right: \emph{semanticity}. Spearman $\rho$ and $p$-values shown per panel. Correlations point in the expected direction on both protocols but do not reach significance.}
%     \label{fig:compressibility}
% \end{figure}

\clearpage
\section{Additional results on alignment with human similarity judgments}\label{app:levels-class}

The Levels dataset~\cite{muttenthaler2025nature} provides human similarity judgment at three different levels, by constructing triplets using 3 images from 1 class to 3 different classes. For completeness, in this section we report results from the third triplet category beyond the coarse- and fine-grained ones reported in Section~\ref{sec:alignment}. In this \emph{border-level} category, two images share the same ImageNet class while the third belongs to a different ImageNet class, probing a granularity that sits between fine and coarse similarity.

The Spearman correlations with border-level alignment fall between the coarse and fine-grained results, with positive but non-significant trends across all three measures: \emph{localizability} ($\rho = 0.58$, $p = 0.223$), \emph{plausibility} ($\rho = 0.72$, $p = 0.110$), and \emph{agreement} ($\rho = 0.42$, $p = 0.402$). The pattern is consistent with interpretability tracking categorical alignment more closely than within-category alignment.

% \begin{figure}[ht!]
%     \centering
%     \includegraphics[width=\linewidth]{figures/class_border.png}
%     \caption{\textbf{Border-level alignment on the Levels dataset.} Each panel plots an interpretability score against odd-one-out accuracy on the border-level triplets of Levels~\cite{muttenthaler2025nature}. Left: \emph{localizability}. Right: \emph{semanticity}. Spearman $\rho$ and $p$-values per panel. Both correlations are positive but non-significant, sitting between the coarse- and fine-grained results.}
%     \label{fig:levels-class}
% \end{figure}

\end{document}